\documentclass[letterpaper, 10 pt, journal, twoside]{ieeetran}
\usepackage{amsmath}
\usepackage{amssymb}
\usepackage[table,xcdraw]{xcolor}
\usepackage{multirow}
\usepackage[utf8]{inputenc}
\usepackage{algorithm}
\usepackage{algorithmic}
\usepackage{graphicx}
\usepackage{subcaption}
\usepackage{hyperref}
\usepackage[capitalise]{cleveref}
\usepackage{caption}
\usepackage[noadjust]{cite}
\usepackage{todonotes}
\usepackage{lscape}

\definecolor{dmfd-green}{RGB}{109,169,85}
\definecolor{dmfd-blue}{RGB}{75,163,253}
\definecolor{dmfd-red}{RGB}{211,42,11}

\newcommand{\gautam}[1]{{\color{red} \textrm{}}}
\newcommand{\gaurav}[1]{{\color{red} \textrm{}}}
\newcommand{\marcus}[1]{{\color{gray} \textrm{}}}
\newcommand{\arthur}[1]{{\color{green} \textrm{}}}
\newcommand{\para}[1]{{\color{blue}\textbf{}}}
\newcommand{\resubmit}[1]{{\color{black} \textrm{#1}}}
\newcommand{\cameraready}[1]{{\color{black} \textrm{#1}}}
\renewcommand{\algorithmiccomment}[1]{\hfill\footnotesize{\textcolor{gray}{$\triangleright$ #1}}}

\newcommand{\argmax}{\arg\!\max} 
 
\newcommand\norm[1]{\| #1 \|}

\newcommand{\buffer}{\mathcal{B}}
\newcommand{\expertTraj}{\mathcal{E}}
\newcommand{\rsi}{\eta}

\newcommand{\state}{\boldsymbol{s}}
\newcommand{\act}{\boldsymbol{a}}
\newcommand{\obs}{\boldsymbol{o}}

\newcommand{\thAct}{\theta}
\newcommand{\actor}{\pi_{\thAct}}
\newcommand{\thCri}{\phi}
\newcommand{\critic}{Q_{\thCri}}
\newcommand{\actPol}{a_{\pi}}
\newcommand{\actBuff}{a_{\buffer}}
\newcommand{\cEnt}{\alpha} %

\newcommand{\traj}{\tau}
\newcommand{\taskDist}{\mathcal{V}}

\usepackage[top=54pt, left=54pt, right=54pt, bottom=54pt]{geometry}

\usepackage{subfiles}

\begin{document}

\title{Learning Deformable Object Manipulation from Expert Demonstrations}
\author{Gautam Salhotra\textsuperscript{*}, I-Chun Arthur Liu\textsuperscript{*}, Marcus Dominguez-Kuhne, \& Gaurav S. Sukhatme\textsuperscript{\textdaggerdbl}\\University of Southern California
\thanks{Gautam Salhotra, I-Chun Arthur Liu, Marcus Dominguez-Kuhne, and Gaurav S. Sukhatme are with the Department of Computer Science, University of Southern California, Los Angeles, CA 90089 USA (e-mail: salhotra@usc.edu; ichunliu@usc.edu; marcusdo@usc.edu; gaurav@usc.edu).}
\thanks{\textsuperscript{*} Equal contribution, \textsuperscript{\textdaggerdbl} G.S. Sukhatme holds concurrent appointments as a Professor at USC and as an Amazon Scholar. This paper describes work performed at USC and is not associated with Amazon.}
}

\markboth{IEEE Robotics and Automation Letters. Preprint Version. Accepted June, 2022}
{Salhotra \MakeLowercase{\textit{et al.}}: Learning Deformable Object Manipulation from Expert Demonstrations}  

\maketitle

\begin{abstract}
We present a novel \resubmit{Learning from Demonstration (LfD)} method, Deformable Manipulation from Demonstrations (DMfD), to solve deformable manipulation tasks using states or images as inputs, given expert demonstrations. 
Our method uses demonstrations in three different ways, and balances the trade-off between exploring the environment online and using guidance from experts \resubmit{to explore high dimensional spaces effectively}.
We test DMfD on a set of representative manipulation tasks for a 1-dimensional rope and a 2-dimensional cloth from the SoftGym suite of tasks, each with state and image observations.
\resubmit{Our method exceeds baseline performance by up to 12.9\% for state-based tasks and up to 33.44\% on image-based tasks, with comparable or better robustness to randomness.}
Additionally, we create two challenging environments for folding a 2D cloth using image-based observations, and set a performance benchmark for them. \resubmit{We deploy DMfD on a real robot with a minimal loss in \cameraready{normalized} performance during real-world execution compared to simulation ($\sim 6\%$).}
\cameraready{Source code is on \href{https://github.com/uscresl/dmfd}{github.com/uscresl/dmfd}.}
\end{abstract}


\section{Introduction}
\label{sec:intro}
\IEEEPARstart{M}{anipulating} deformable objects is a formidable challenge: extracting state information and modeling are both difficult problems and the task is dauntingly high-dimensional with a large action space. {\em Learning} to manipulate deformable objects from expert demonstrations may offer a way forward to alleviate some of these problems.

\begin{figure}[th!]
\begin{subfigure}{\columnwidth}
  \centering
  \includegraphics[width=\linewidth]{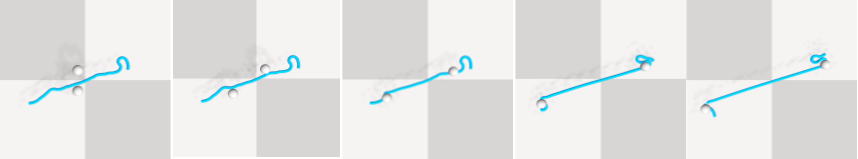}
  \caption{Straighten Rope}
  \label{fig:rollout-rope-straighten}
\end{subfigure}
\begin{subfigure}{\columnwidth}
  \centering
  \includegraphics[width=\linewidth]{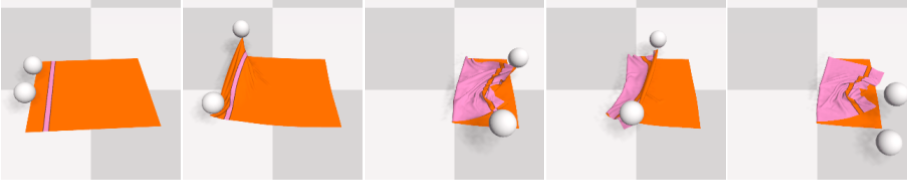}
  \caption{Cloth Fold}
  \label{fig:rollout-cf}
\end{subfigure}
\begin{subfigure}{\columnwidth}
  \centering
  \includegraphics[width=\linewidth]{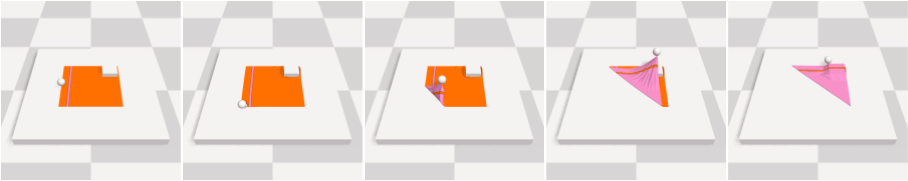}
  \caption{Cloth Fold Diagonal Pinned}
  \label{fig:rollout-cfd-pinned}
\end{subfigure}
\begin{subfigure}{\columnwidth}
  \centering
  \includegraphics[width=\linewidth]{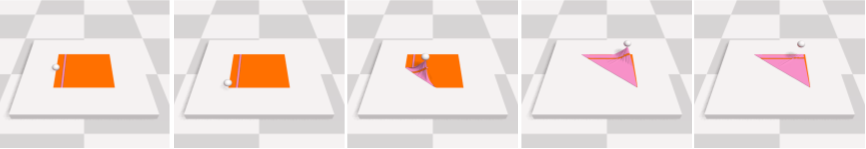}
  \caption{Cloth Fold Diagonal Unpinned}
  \label{fig:rollout-cfd-unpinned}
\end{subfigure}
\begin{subfigure}{\columnwidth}
  \centering
  \includegraphics[width=\linewidth]{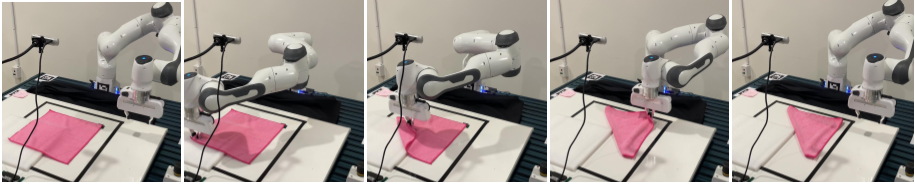}
  \caption{Cloth Fold Diagonal Unpinned on real robot}
  \label{fig:rollout-cfd-unpinned-real}
\end{subfigure}
\caption{
\textbf{Learning deformable manipulation} For our method DMfD, we describe a learned agent that achieves state-of-the-art performance \resubmit{among methods that use expert demonstrations}, for solving difficult \resubmit{deformable} manipulation tasks such as straightening 1D ropes and folding 2D cloths based on scene images. 
\resubmit{ We set a new benchmark} on the Straighten Rope~(\autoref{fig:rollout-rope-straighten}) task which requires the agent to straighten a rope with two end effectors, shown as white spheres, and on the Cloth Fold~(\autoref{fig:rollout-cf}) task which requires the agent to fold a flattened cloth into half, along an edge.
Both tasks are from the SoftGym suite~\cite{Lin2020SoftGym}. 
Additionally, we introduce and solve a new task constrained to a single end effector - the Cloth Fold Diagonal task, 
which requires an agent to fold a square cloth along a diagonal. 
In the pinned version~(\autoref{fig:rollout-cfd-pinned}) of this task, the cloth is clamped to the table at a corner; in the unpinned version~(\autoref{fig:rollout-cfd-unpinned}) it is  not. 
\resubmit{\autoref{fig:rollout-cfd-unpinned-real} shows the unpinned version being executed on a real robot.}}
\label{fig:hero-figure}
\end{figure}

\resubmit{We present a new Learning from Demonstration (LfD) method -- Deformable Manipulation from Demonstrations (DMfD) -- that works with high-dimensional state or image observations. It absorbs expert guidance, whether from human execution or hand-engineered, while learning online to solve challenging deformable manipulation tasks such as cloth folding. DMfD is an asymmetric actor-critic method that uses expert data in three ways: 1. the replay buffer is pre-populated with expert trajectories before training, 2. during training, we leverage an advantage-weighted loss, where the replay buffer samples are weighted to encourage the policy to stay close to the stored expert actions, and 3. during experience collection using reference state initialization. Our results show that non-trivial and novel combination of these equips the agent with the ability to explore high dimensional spaces effectively while leveraging guidance from expert demonstrations. Our contributions are as follows.}
\resubmit{
\begin{enumerate}
    \item To encourage wide exploration, we {\em add  an exploration term (a soft state value function)} to the advantage-weighted loss. This term samples actions according to the current policy instead of actions from the replay buffer. This is an improvement over the original advantaged-weighed formulation~\cite{peng2019AWR,nair2020AWAC} which only samples actions from the replay buffer to update its policy. To deploy our methods in real-world settings, we extend the advantage-weighted framework to the image domain, using CNNs and data augmentation (random crops~\cite{Kostrikov2020DrQ}) to prevent overfitting.
    \item During experience collection, we {\em introduce  probabilistic} reference state initialization (RSI). Instead of  always resetting the agent to the states seen by the expert~\cite{Peng2018DeepMimic}, we invoke RSI probabilistically. This promotes exploration and learning in states that are difficult to reach (for example, due to high dimensionality or the dynamics of the environment) while the agent has the opportunity to learn from previously seen states. 
    \item We {\em create two new environments} (2D deformables, image-based observations), with one robot arm. We {\em deploy DMfD on a real robot} with a minimal sim2real gap ($\sim$6\%), indicating that it can work in real-world settings.
    \item DMfD outperforms \resubmit{LfD and non-LfD} baselines on both state-based environments (by up to 12.9\% median performance) and on image-based environments (by up to 33.44\% median performance). Sample rollouts of our method for difficult image-based manipulation tasks \resubmit{and real robot experiments} can be seen in \autoref{fig:hero-figure}.
\end{enumerate} 
}
\color{black}

\section{\resubmit{Background}}
\label{sec:background}

Deformable object manipulation has been a challenge in robotics with many real-world applications, such as folding clothes~\cite{Yilin2019DeformNoDemos}, cooking food~\cite{kolathaya2018direct}, or assisting humans~\cite{joshi2017robotic}. 
Its high-dimensional state representation and complex dynamics make manipulation tasks significantly more difficult than rigid body manipulation. Traditionally, analytical methods have been employed to solve deformable object manipulation tasks. 
Methods such as Finite Element Method~\cite{wriggers2008FEMBook} are used to model object dynamics. 
Control methods such as trajectory optimization~\cite{zimmermann2021dynamic} and model predictive control~\cite{allgower2012nonlinearMPC} use these models to specify control inputs and manipulate the object. 
Although these have proven to be successful under certain conditions, it is difficult to generalize them to perturbations or variations in the environment. Recently, data-driven methods have gained popularity in solving manipulation tasks~\cite{Preiss2022TrajsDeform}, such as Imitation Learning (IL) \cite{Laskey2017Dart, Torabi2018BehavorialCloning, Ho2016GAIL},  Reinforcement Learning (RL) \cite{danielczuk2019mechanicalsearch, Kurenkov2019ACTeach, Zeng2018PushGraspSyn, yamada2020mopa}, and combining IL with RL \cite{Horgan2018DistPriExpReplay, liu2021mopa}. 
However, most of the successes have been in rigid body manipulation. 
The low observability and controllability of deformable objects, coupled with the typically high dimensionality of the parameter space in learning methods, make it challenging for learning alone to solve these tasks. 
Here, we focus on deformable object manipulation with a novel expert-guided RL method. 

\noindent{\bf Learning from Expert Demonstrations:} Two common methods of learning from demonstrations include IL and Offline RL. 
IL is a powerful machine learning technique used to imitate expert demonstrations. 
IL has been applied to soft body manipulation e.g., DART~\cite{seita2019bedmaking} has been used for bed making, where human demonstrations are used on a robot and the Transporter Network~\cite{Seita2021DeformableRavens} with goal conditioning has been used for manipulating beads, cloths, and bags. 
Dynamic Movement Primitives have been used~\cite{joshi2017robotic} to learn cloth manipulation from demonstrations. 
The common issue with these methods is that they tend to fail when encountering a new state due to the accumulating errors from covariate shift \cameraready{\cite{ross2010efficient}}. 
Moreover, these methods' performance is often bounded by the quality of expert demonstrations.
Similar to IL, Offline RL~\cite{lange2012batch, levine2020OfflineRL, rashidinejad2021bridging} generally learns from past demonstrations without online environment interactions. 
Offline RL has two properties: all transitions are stored in an offline dataset, and network updates occur on the entire batch of transitions. In particular, Offline RL can handle large, diverse datasets which produce more generalizable policies~\cite{rashidinejad2021bridging}. But they often achieve sub-optimal performance when used in online fine-tuning, discussed in~\cite{nair2020AWAC}.

\noindent{\bf Reinforcement Learning (RL):} Reinforcement learning enables an agent to learn in an interactive environment via trial and error.
RL has been applied to manipulation problems~\cite{joshi2017robotic, rajeswaran2017learningManip}; additionally \cite{yamada2020mopa} empowers RL agents with motion planning techniques to manipulate cubes and assemble furniture and \cite{liu2021mopa} extends it to the visual domain. However, most of these apply to rigid body manipulation. 
A limited number of RL methods have been used for deformable manipulation~\cite{Lin2020SoftGym,matas2018sim, Yilin2019DeformNoDemos}, some e.g., CURL~\cite{Laskin2020CuRL} and DrQ~\cite{Kostrikov2020DrQ} using vision.
\cite{nguyen2019reviewOfManip} provides a thorough overview of reinforcement learning techniques used for robotic manipulation tasks.

\noindent{\bf Combining Reinforcement Learning and Expert Guidance:}
IL techniques are trained to perform a task from demonstrations by learning the mapping between observations and actions.
Hence, when demonstrations can be easily given for a problem, IL is a preferred method. 
RL, in contrast, is suitable when a reward function can be easily specified and the environment can be easily explored.
However, it is time-consuming to naively explore the state space without expert demonstrations. 
Thus, there have been several studies \cite{Horgan2018DistPriExpReplay, nair2018LfDExploration} \cameraready{\cite{zhureinforcementRSS2018}} focusing on how to combine IL and RL effectively gaining the advantages of IL, where the agent explores by learning from expert demonstrations, and RL, where the agent learns to improve the policy further. 
Deep Mimic~\cite{Peng2018DeepMimic} uses RSI to address exploration cost by initializing from past high-value states, since some high reward states may be difficult to reach but valuable for exploration. 
Advantage Weighted Actor Critic~(AWAC)~\cite{nair2020AWAC} is another method that utilizes expert demonstrations. 
It proposes an implicit policy constraint to efficiently train an off-policy RL algorithm to learn from offline data followed by online fine-tuning.
Although these methods are not specifically designed for deformable object manipulation, they have shown significant performance improvements in other areas. 
In this work, we demonstrate that RL combined with appropriate use of expert data can greatly improve deformable object manipulation.

\section{Formulation and Approach}
\resubmit{\label{sec:formulation}}

We formulate the deformable manipulation problem as a partially observable Markov decision process (POMDP).
Consider a POMDP with state space $\mathcal{S}$, action space $\mathcal{A}$, observation space $\mathcal{O}$, discount factor $\gamma$, horizon $H$, dynamics function $\mathcal{T}: \mathcal{S} \times \mathcal{A} \rightarrow \mathcal{S}$ and reward function $r: \mathcal{S} \times \mathcal{A} \rightarrow \mathbb{R}$. 
At time $t$, the agent is at state $\state\in \mathcal{S}$, gets an observation $\obs \in \mathcal{O}$ and takes action $\act \in \mathcal{A}$.
It reaches state $\state'$, and gets back an observation $\obs'$, reward $r_t = r(\state_t, \act_t)$.
The discounted reward from time $t$ is given by $R_t = \sum_{i=t}^H \gamma^i r_i$. We generalize a single task over a family of variants $\taskDist$ that determine properties of the object to be manipulated.
The initial state is a function of the variant, $\state_0(v), v \sim \taskDist$.

\begin{figure}[!t]
    \centering
    \includegraphics[width=\columnwidth]{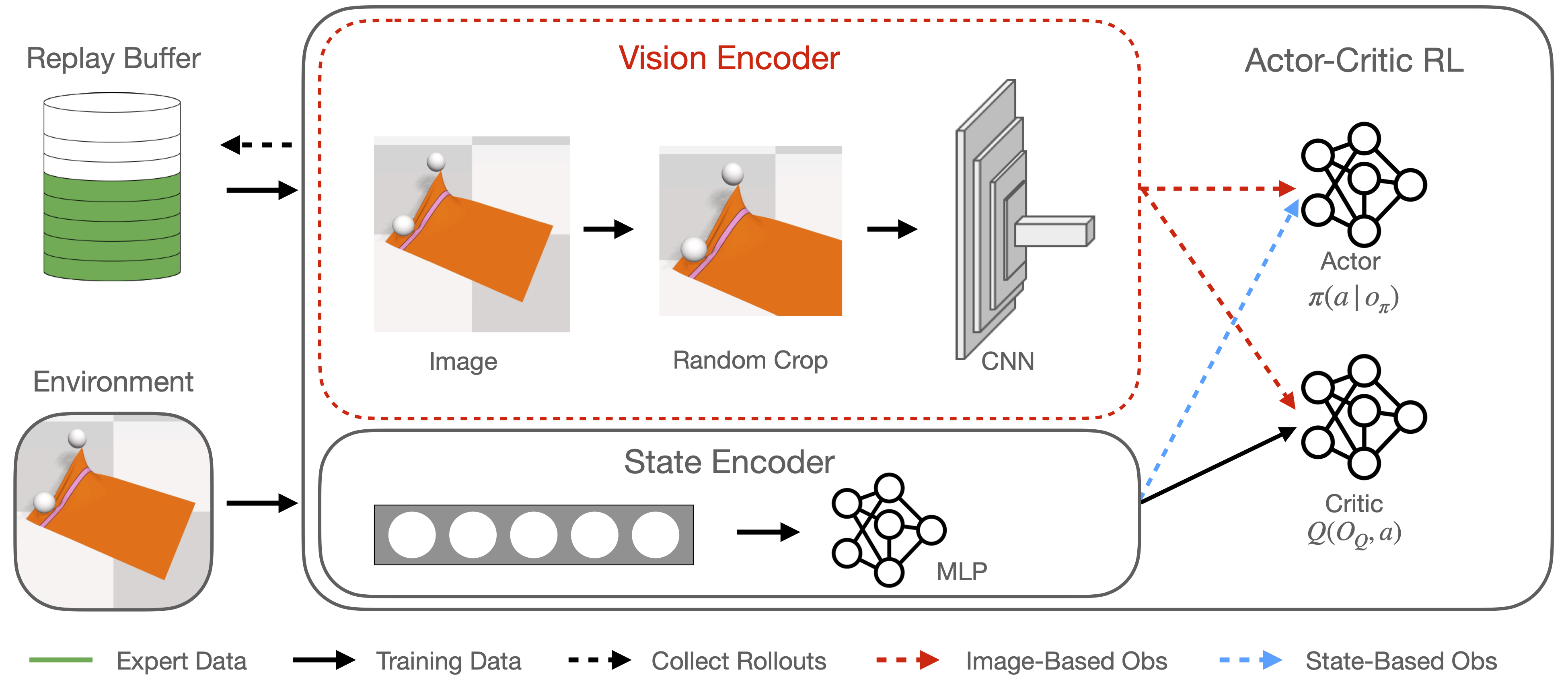}
    \caption{
    \textbf{Schematic of our method.} The agent obtains observations from the environment (during experience collection) or the replay buffer $\buffer$ (during training). Pre-populated expert demonstrations in the replay buffer are shown in \textcolor{dmfd-green}{\textbf{Green}}.  
    The training pipeline works with state-based or image-based observations. 
    With state-based observations, the actor and critic get an encoding of the system state~($o_Q = o_\pi = o_{\state}$), shown as \textbf{Black} and \textcolor{dmfd-blue}{\textbf{Blue}} arrows. 
    With image-based observations, the actor gets an encoding of the image whereas the critic gets encodings of the both the state and the image ($o_\pi = o_{image}, o_Q = o_{\state} \cup o_{image}$), denoted by \textbf{Black} and \textcolor{dmfd-red}{\textbf{Red}} arrows.
    }
    \label{fig:schematic_unified}
\end{figure}

The problem reduces to finding the best policy  $\pi \in \Pi$, that maximizes the expected discounted reward $J(\pi)$ of an episode, over task variants $v$ and the distribution induced by the policy.
\begin{equation}
    J(\pi) = \mathop{\mathbb{E}}_{\tau \sim \pi(\tau), v \sim \mathcal{V}} [R_{0}]
\end{equation}

subject to $\state_{t+1} = \mathcal{T}(\state_t,\act_t)$, and initial state $\state_0(v)$.
$\pi(\traj)$ is the likelihood of trajectory $\tau=(\state_0,\act_0,\state_1,\act_1,\dots,\state_H)$ under policy $\pi$ and initial condition $\state_0$.

We assume the availability of expert data, which may be hand-engineered solutions, demonstrations by a human expert, or any other method of procuring trajectories that solve the task.
Thus, we have a demonstration dataset that we wish to learn from, in addition to the agent's rollouts during experience collection.
We choose to maximize expected advantage $A^\pi(s_t,a_t)$ instead of the return $R_t$ because it is an unbiased estimator of the expected return with lower variance~\cite{williams1992simple}.
We maximize this advantage over a sampling of transitions from a replay buffer $\buffer$ of a mixture of policies, using a sampling policy $\pi_\buffer$.  
This formulation is similar to Advantage-Weighted Regression (AWR)~\cite{peng2019AWR} with experience replay over a mixture of policies.
Our policy optimization problem can be defined as maximizing advantage while remaining close to the sampling policy.

\begin{equation}
    \pi^* = \mathop{\argmax}_{\pi \sim \Pi} \mathop{\mathbb{E}}_{\state \sim d_\pi(\state)} \mathop{\mathbb{E}}_{\act \sim \pi(\cdot |\state)} [A^{\pi}(\state,\act)]
\end{equation}

\begin{equation}
    s.t.\ \ D_{KL}(\pi(\cdot | \state) \| \pi_{\buffer} (\cdot | \state) ) \leq \epsilon
\end{equation}
where $d_\pi(\state)$ is a state distribution induced by $\pi$ and $D_{KL}$ is the KL divergence.
Following AWR, we reduce the objective and constraint to an advantage-weighted objective for a policy with parameters~$\thAct$
\begin{equation}
    \mathcal{L}_{A} = \mathop{\mathbb{E}}_{\state,\act \sim \buffer} \left[\log{\actor(\act|\state)} \exp{\left( \frac{1}{\lambda} A^{\pi}(\state,\act) \right)} \right]
    \label{eq:awr_awac_loss}
\end{equation}
where $\lambda$ is a temperature parameter (see~\cite{peng2019AWR} for a complete derivation).
The loss function $\mathcal{L}_Q$ for the critic $Q_\phi$ (with parameters $\phi$) is based on the error between the estimated Q-value $q_{\phi, B}$ and the Bellman update $b$.
\begin{equation}
    \mathcal{L}_{Q} = \mathbb{E}_\buffer[\norm{q_{\thCri,\buffer} - b}^2]
    \label{eq:critic_loss}
\end{equation}
where $b = r + \gamma\mathbb{E}[Q_{\phi}(\state ',\act ')]$ during the episode and $b = r$ at the last timestep $t=H$. Since state estimation is difficult for deformable manipulation, we extend this formulation to the partially observable case. 
Thus, the policy acts on the observation $\actor(\act|\obs)$ instead of the state $\actor(\act|\state)$.

\begin{algorithm}[t]
    \caption{Deformable Manipulation from Demonstrations}
    \label{alg:method}
    \begin{algorithmic}[1]
        \renewcommand{\COMMENT}[1]{{\algorithmiccomment{#1}}}
        \REQUIRE Task distribution $\taskDist$, Expert trajectories $\expertTraj$
        \STATE Initialize replay buffer $\buffer = \expertTraj$
        \STATE Initialize $\actor, \critic$

        \FOR {iteration $i = 1, 2, \dots$}
            \STATE Sample batch $(\state,\obs,\boldsymbol{\actBuff},\obs',r,d) \sim \buffer$
            \STATE Get current policy $\boldsymbol{\actPol} \sim \pi_{\thAct}(o)$
            \STATE Compute critic loss $\mathcal{L}_Q$ as in \autoref{eq:critic_loss}

            \STATE $\thCri \leftarrow OPT(\thCri, \bigtriangledown \mathcal{L}_Q)$
            \COMMENT{Optimize critic}
            \STATE Compute actor loss $\mathcal{L}_{\pi}$ as in \autoref{eq:actor_loss}
            \STATE $\thAct \leftarrow OPT(\thAct, \bigtriangledown \mathcal{L}_\pi)$
            \COMMENT{Optimize actor}
            \STATE$\traj_1, \traj_2, ..., \traj_K \sim \actor(\traj)$
            \COMMENT{Experience collection}
            \STATE $\buffer \leftarrow \buffer \cup \{\traj_1, \traj_2, ..., \traj_K\}$
        \ENDFOR
        \STATE Return $\actor$
    \end{algorithmic}
\end{algorithm}

Our (actor-critic) method learns from an expert dataset, while having access to online interaction with the environment. 
Before training, we populate the replay buffer $\buffer$ with expert trajectories $\expertTraj$, (replay-buffer spiking~\cite{lipton2018bbq_populatebuffer}). 
This is known to improve performance (even with few episodes), since it shows the existence of a good policy with large reward.
It helps the algorithm realize good actions early on (\autoref{sec:abl-prepopulatebuffer}).
Unlike offline RL, we have easy access to the simulator giving the agent the ability to explore the environment to find potentially better trajectories than the offline expert dataset. 
To promote this, we update the replay buffer during training, thus updating the mixture of policies that make up the sampling policy. 
Thus, we have the ability to learn from, and even exceed, expert data in the environment. We add entropy regularization to the actor, to balance exploration and exploitation \cite{Toumas2018SAC}. 
We require that the policy maximize an entropy-regularized version of the value function
\begin{equation}
    V(\state) = \mathop{\mathbb{E}}_{\act\sim\pi(\cdot|\state)} [Q(\state,\act) - \cEnt\log{\pi_\thAct}(\act|\obs)]
\end{equation}
where $\alpha$ is a weighting hyper-parameter and $\act$ is sampled from the current policy.
We propose an entropy loss term to minimize,
\begin{equation}
    \mathcal{L}_{E} = \mathop{\mathbb{E}}_{\state,\act,\obs \sim \buffer}[\cEnt \log{\actor(\act|\obs)} - Q(\state,\act)]
    \label{eq:entropy_loss}
\end{equation}

Our policy loss is $w_{E}$-weighted linear combination,
\begin{equation}
    \mathcal{L}_\pi = (1 - w_{E})\mathcal{L}_A + w_{E} \mathcal{L}_{E},\ 0 \leq w_{E} \leq 1
    \label{eq:actor_loss}
\end{equation}

While this does not have a tractable closed-form solution, we can optimize it numerically with gradient steps. As is typical, we alternate gradient steps for actor and critic respectively.
The algorithm is shown in Alg. \autoref{alg:method}.

During experience collection, with a tuned probability $p_{\rsi}$, we reset the robot to some environment state that the expert was in. 
We then compare the trajectory of the agent with the trajectory of the expert and provide an imitation reward based on the states achieved.
Reference state initialisation~(RSI)~\cite{Peng2018DeepMimic} was introduced for dynamic tasks. 
It helps to explore and learn in high-dimensional states that are difficult to reach. 
However, always using RSI (i.e., always reset to a state the expert has seen) prevents the agent from exploring the environment freely and may lead to overfitting to those demonstrations. 
As \autoref{sec:abl-rsi} discusses, both 0\% and 100\% RSI are worse than probabilistic RSI, implying that expert guidance helps when applied sparingly. 
Thus, once again, we have the ability to learn from and exceed the expert. 
Probabilistic RSI is similar to replay buffer spiking~\cite{lipton2018bbq_populatebuffer} referenced above, in that knowing the existence of \textit{some} good actions and rewards (without using \textit{only} those) is beneficial. 
Further, this decreased dependence on experts allows us to work with suboptimal experts, potentially reducing the burden on the human expert. 

\cameraready{Our state encoder network} is composed of multi-layer perceptrons with $\tanh$ activation, as we normalize our actions to $[-1,1]$. 
\cameraready{Our image encoder network is} a Convolutional Neural Network (CNN) to process images.
We also augment the input image with random crops, a known improvement for vision-based reinforcement learning~\cite{Kostrikov2020DrQ}
\autoref{fig:schematic_unified} shows these architectures. 
Note that the critic receives state input in addition to the observation, whereas the actor only gets the type of observation chosen for the environment. 
This asymmetry has been shown to be useful for stabilising the critic~\cite{pinto2017A2C}, and is justified in \autoref{sec:abl-critic}.
Network specifics are given in \autoref{sec:experiments}. 

\color{black}

\section{Experiments}
\label{sec:experiments}

\begin{figure*}%
\begin{subfigure}{0.7\columnwidth}
  \centering
  \includegraphics[width=\linewidth]{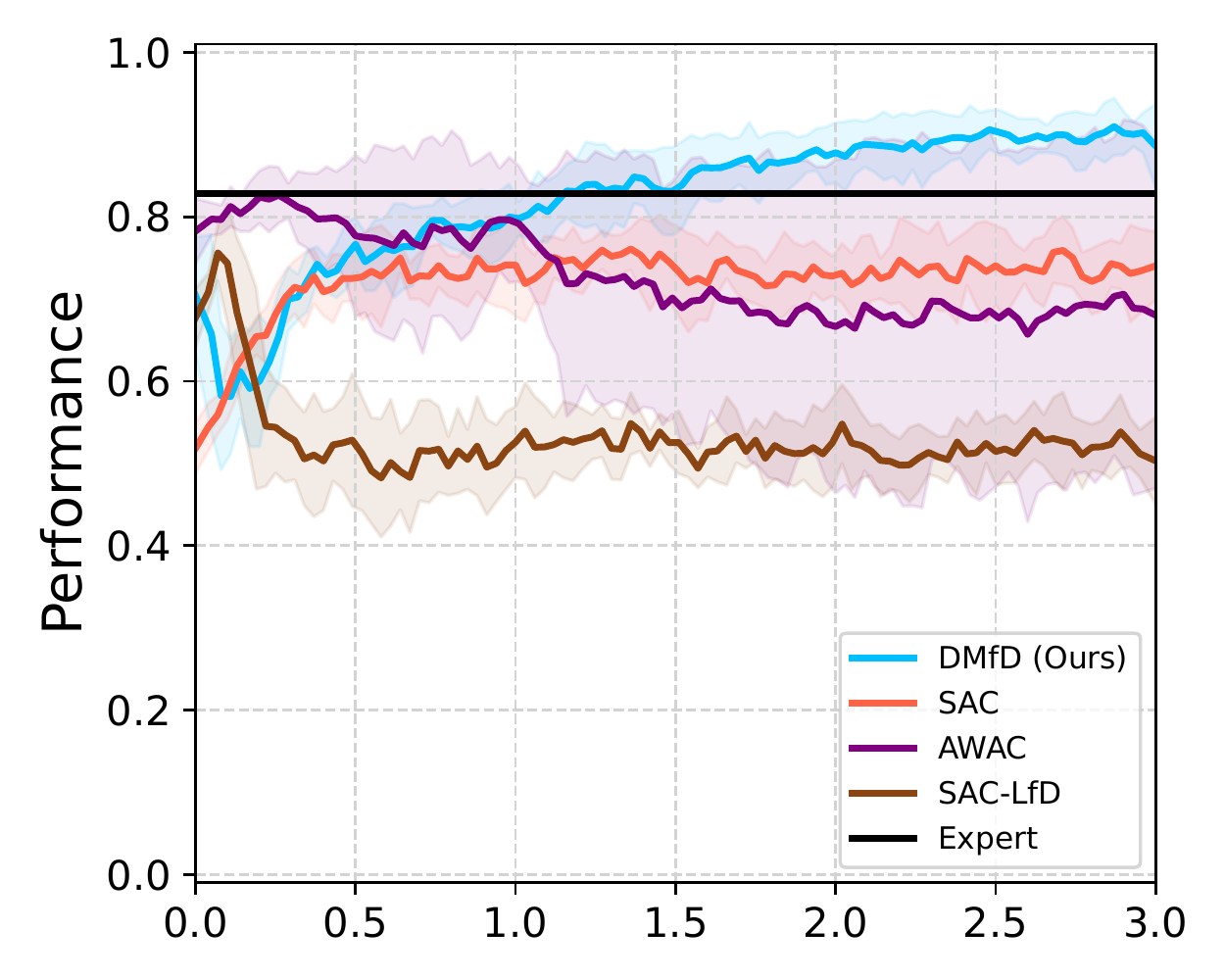}
  \caption{\resubmit{Straighten Rope State}}
  \label{fig:sota-rope-straighten-state}
\end{subfigure}
\begin{subfigure}{0.7\columnwidth}
  \centering
  \includegraphics[width=\linewidth]{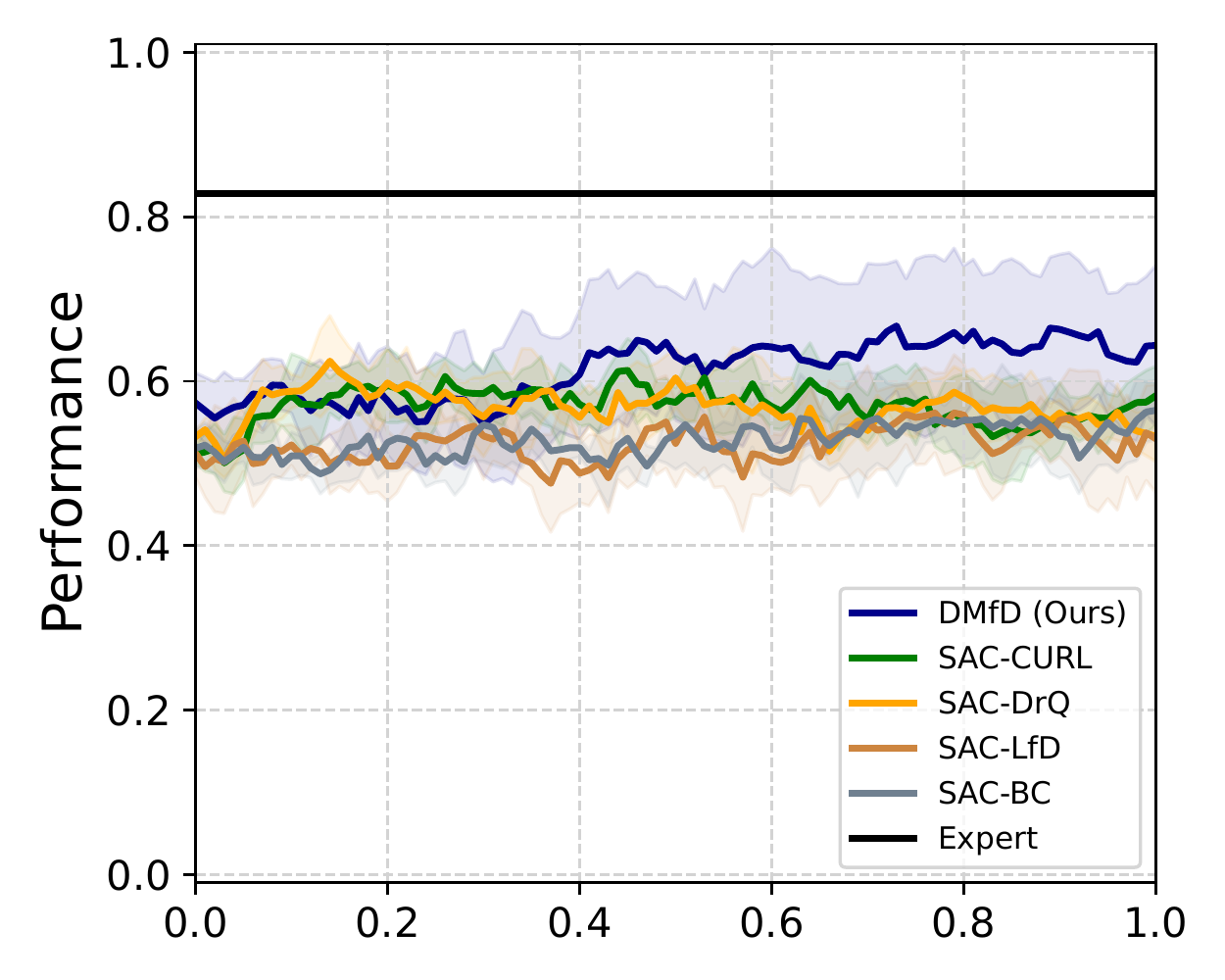}
  \caption{\resubmit{Straighten Rope Image}}
  \label{fig:sota-rope-straighten-image}
\end{subfigure}
\begin{subfigure}{0.7\columnwidth}
  \centering
  \includegraphics[width=\linewidth]{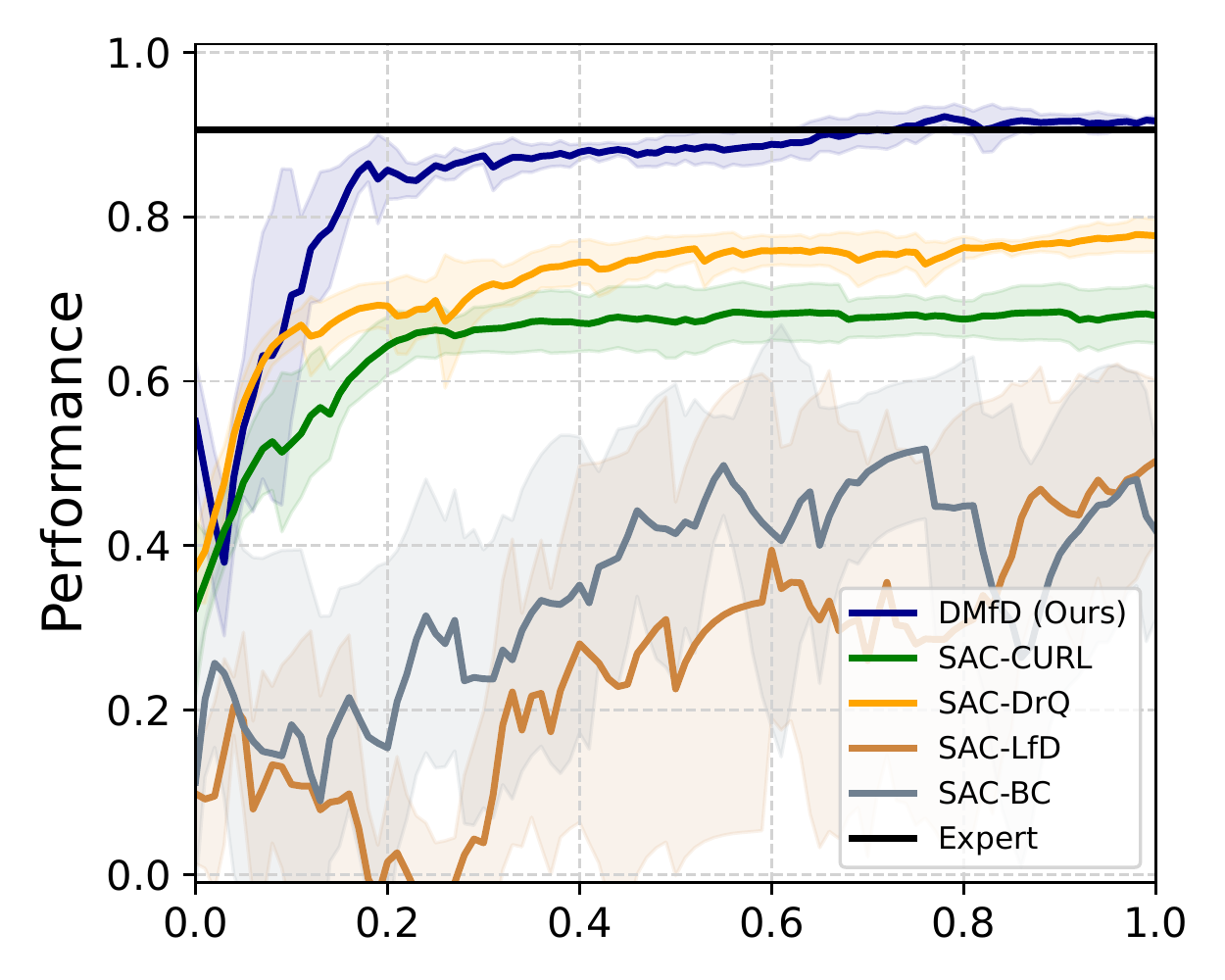}
  \caption{\resubmit{Cloth Fold Diagonal Pinned Image}}
  \label{fig:sota-cloth-fold-diagonal-pinned-image}
\end{subfigure}
\begin{subfigure}{0.7\columnwidth}
  \centering
  \includegraphics[width=\linewidth]{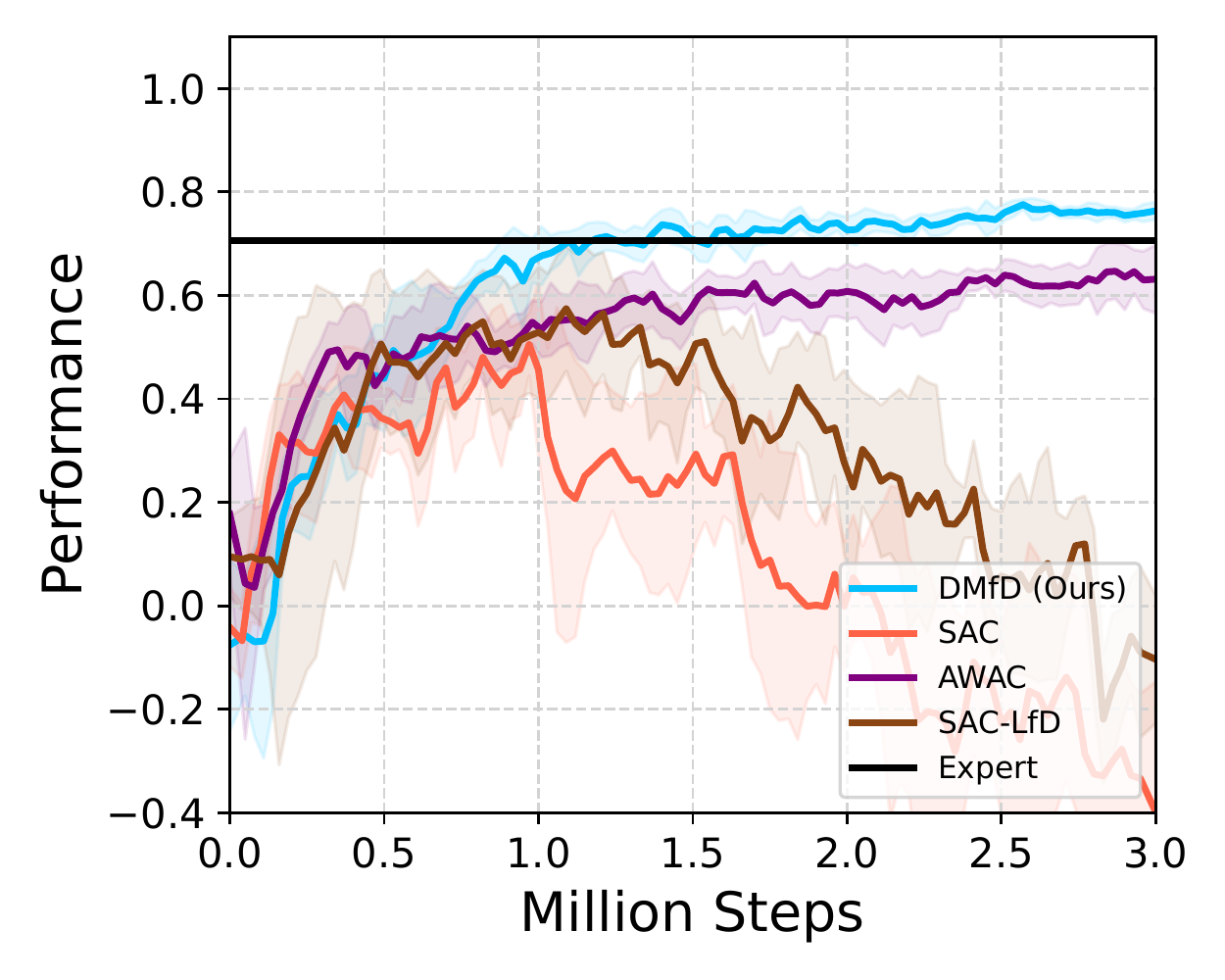}
  \caption{\resubmit{Cloth Fold State}}
  \label{fig:sota-cloth-fold-state}
\end{subfigure}
\begin{subfigure}{0.7\columnwidth}
  \centering
  \includegraphics[width=\linewidth]{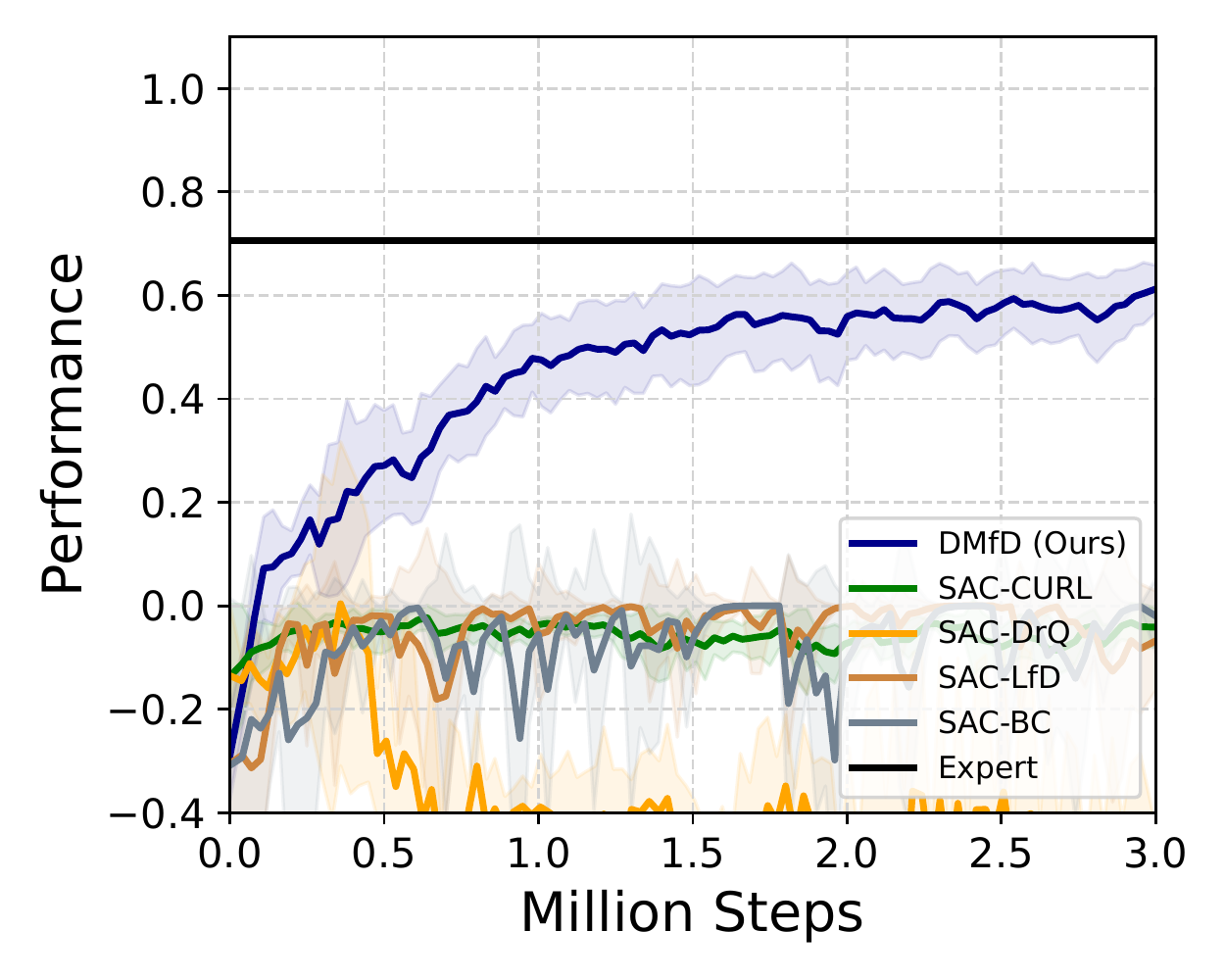}
  \caption{\resubmit{Cloth Fold Image}}
  \label{fig:sota-cloth-fold-image}
\end{subfigure}
\begin{subfigure}{0.7\columnwidth}
  \centering
  \includegraphics[width=\linewidth]{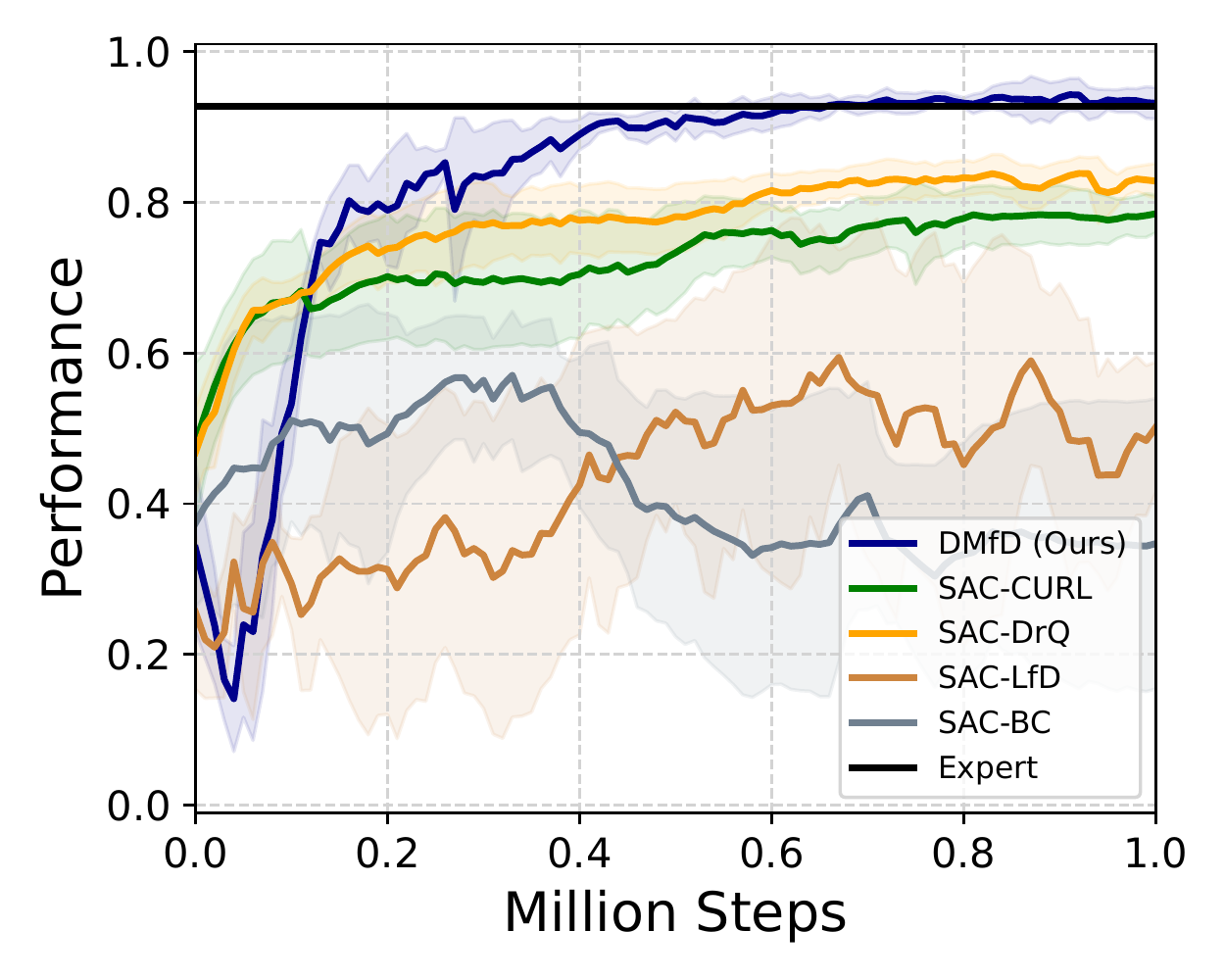}
  \caption{\resubmit{Cloth Fold Diagonal Unpinned Image}}
  \label{fig:sota-cloth-fold-diagonal-unpinned-image}
\end{subfigure}
\caption{\textbf{\resubmit{Performance} comparisons.} Learning curves of the normalized performance $\hat{p}(H)$ for all environments during training. The first column(\ref{fig:sota-rope-straighten-state} \& \ref{fig:sota-cloth-fold-state}) shows SoftGym state-based environments. The second column(\ref{fig:sota-rope-straighten-image} \& \ref{fig:sota-cloth-fold-image}) shows SoftGym image-based environments, and the third column (\ref{fig:sota-cloth-fold-diagonal-pinned-image} \& \ref{fig:sota-cloth-fold-diagonal-unpinned-image}) shows our new Cloth Fold Diagonal environments. 
\resubmit{All environments were trained until convergence.}
State-based DMfD is in light blue, and the image-based agent is in dark blue. 
The expert performance is the solid black line. 
We compare against the baselines described in~\autoref{sec:sota-comparison}. 
Behavioural Cloning does not train online, its results are shown in \autoref{tab:sota-comparisons-all}.
We plot the mean $\mu$ of the curves as a solid line, and shade one standard deviation ($\mu \pm \sigma$). DMfD consistently beats the baselines, with comparable or better variance. 
For a detailed discussion see~\autoref{sec:discussion}.
}
\label{fig:sota-comparisons-all}
\end{figure*}

\subsection{Tasks and Experimental Setup}
\label{sec:task-descriptions}
We use four different tasks and two different observation types in our experiments (below), all of which are conducted in the SoftGym suite~\cite{Lin2020SoftGym}.
We encode object states with an object-specific reduced-state that SoftGym provides, and use it to train all methods that require object state as input. Details for the reduced-state representation for each task are given below.
The image observation is a 32x32 RGB image of the environment showing the object and robot end-effector.
Each task has a number of deformable object property variants for effective domain randomization. 

    \noindent \textbf{Straighten Rope:} 
    The objective is to stretch the ends of the rope a fixed distance from each other, to force the rope to be straightened. 
    The reduced state is the $(x,y,z)$ coordinates of 10 equidistant keypoints along the rope, including the endpoints. 
    \resubmit{Performance is measured by comparing the distance between endpoints to a fixed length parameter.}

    \noindent \textbf{Cloth Fold:} 
    The objective is to fold a flattened cloth into half, along an edge, using two end-effectors.
    The reduced state is the $(x,y,z)$ coordinates of each corner. 
    \resubmit{Performance is measured by comparing how close the left and right corners are to each other.}

    \noindent \textbf{Cloth Fold Diagonal Pinned:} 
    The objective is to fold the cloth along a specified diagonal of a square cloth, with a single end-effector. 
    One corner of the cloth is pinned to the table by a heavy block.
    The reduced state is the $(x,y,z)$ coordinates of each corner. 
    \resubmit{Performance is measured by comparing how close the bottom-left and top-right corners are to each other.}
    This is a new task introduced in this paper.
    
    \noindent \textbf{Cloth Fold Diagonal Unpinned:} 
    The objective is to fold the cloth along a specified diagonal of a square cloth, with a single end-effector. 
    The cloth is free to move on the table top.
    The reduced state is the $(x,y,z)$ coordinates of each corner. 
    \resubmit{Performance is measured by comparing how close the bottom-left and top-right corners are to each other.}
    This is a new task we introduce in this paper.

In each task, image-based environments were observed to be more difficult to solve than state-based environments; thus for the new Cloth Fold Diagonal tasks we focus on the more difficult (image-based) setting. 
This produces 6 environments (4 from SoftGym: state- and image-based settings for Straighten Rope and Cloth Fold) and 2 newly introduced here (both image-based settings for Cloth Fold Diagonal Pinned and Cloth Fold Diagonal Unpinned). We create demonstrations using hand-engineered solutions, where the expert is an oracle with access to the full state and dynamics. 

The following subsections compare the performance of agents in each task, as measured by a normalized metric (in $[0,1]$) described in SoftGym.
The normalized performance at time $t$, $\hat{p}(t)$ is given by $\hat{p}(t) = (p(\state_t) - p(\state_0))/(p_{opt} - p(\state_0))$
where $p$ is the \resubmit{environment-specific} performance function of state $\state_t$ at time $t$, and $p_{opt}$ is the best possible performance on the task. As in SoftGym, we use the normalized performance at the end of the episode, $\hat{p}(H)$.

We used an actor critic model with the actor and critic networks both having 2 hidden 1024-wide layers with $tanh$ activations. 
Additionally, for vision input, we use a convolutional neural network with 4 convolution layers each with 32 channels, single stride, a 3x3 kernel and LeakyReLU activation functions, followed by 2 1024-wide dense layers. 
Additionally, our RSI probability $p_\rsi$ was 0.2 for state and 0.3 for image observations. 
Our entropy regularization weight was $w_{E}=0.1$, with coefficient $\cEnt=0.5$ in entropy regularization and a discount factor of $\gamma=0.9$. 
Additionally our expert dataset was optimally tuned to hold 8,000 episodes each with an episode horizon of 
\resubmit{75} steps.

We ran our experiments on a server with Intel Xeon CPU cores (3.00GHz) and NVIDIA GeForce RTX 2080 Ti GPUs. Our experiments ran with 16 CPUs and 1 GPU allocated. 
We ran image-based methods for 1M steps, as in SoftGym experiments, but were able to run state-based methods for 3M steps as they were much faster to train, due to the low dimensional reduced-state input. 
With high dimensional inputs such as images, even elements such as the replay buffer, expert dataset, and vision encoder need a lot more memory and compute, which slows down training. 
For example, we observed that when running our experiment on the Cloth Fold environment, it took 34 hours to run 1M steps on the image-based version, and 39 hours to run 3M steps on the state-based version.

\subsection{\resubmit{Performance} Comparisons}
\label{sec:sota-comparison}

\begin{table*}[]
    \centering
    \begin{tabular}{lllllll
    >{\columncolor[HTML]{EFEFEF}}l cllll}
     & \multicolumn{6}{c}{\cellcolor[HTML]{EFEFEF}\textbf{Image-based Environments}} &  & \multicolumn{5}{c}{\cellcolor[HTML]{EFEFEF}\textbf{State-based Environments}} \\ \cline{2-7} \cline{9-13} 
     & \multicolumn{1}{c}{{\textbf{\begin{tabular}[c]{@{}c@{}}BC-\\ image\end{tabular}}}} & {\textbf{SAC-BC}} & {\textbf{SAC-LfD}} & \multicolumn{1}{c|}{\textbf{\begin{tabular}[c]{@{}c@{}}DMfD\\ (ours)\end{tabular}}} & \multicolumn{1}{c}{\textbf{DrQ}} & \multicolumn{1}{c}{\textbf{CURL}} & \multicolumn{1}{c}{\cellcolor[HTML]{EFEFEF}\textbf{\begin{tabular}[c]{@{}c@{}}Expert\\ (state)\end{tabular}}} & \multicolumn{1}{c|}{\textbf{SAC}} & \multicolumn{1}{c}{\textbf{AWAC}} & \multicolumn{1}{c}{{\textbf{\begin{tabular}[c]{@{}c@{}}BC-\\ state\end{tabular}}}} & {\textbf{SAC-LfD}} & \multicolumn{1}{c}{\textbf{\begin{tabular}[c]{@{}c@{}}DMfD\\ (ours)\end{tabular}}} \\ \cline{2-7} \cline{9-13} 
    \multicolumn{1}{c}{\textbf{}} & \multicolumn{6}{c}{\textbf{Straighten Rope Image}} &  & \multicolumn{5}{c}{\textbf{Straighten Rope State}} \\ \cline{2-7} \cline{9-13} 
    $\boldsymbol{\mu \pm \sigma}$ & \begin{tabular}[c]{@{}l@{}}0.454\\ $\pm$0.256\end{tabular} & {\begin{tabular}[c]{@{}l@{}}0.600\\ $\pm$0.085\end{tabular}} & {\begin{tabular}[c]{@{}l@{}}0.519\\ $\pm$0.280\end{tabular}} & \multicolumn{1}{l|}{\textbf{\begin{tabular}[c]{@{}l@{}}0.668\\ $\pm$0.256\end{tabular}}} & \begin{tabular}[c]{@{}l@{}}0.536\\ $\pm$0.259\end{tabular} & \begin{tabular}[c]{@{}l@{}}0.551\\ $\pm$0.262\end{tabular} & \begin{tabular}[c]{@{}l@{}}0.829\\ $\pm$0.099\end{tabular} & \multicolumn{1}{l|}{\begin{tabular}[c]{@{}l@{}}0.701\\ $\pm$0.246\end{tabular}} & \begin{tabular}[c]{@{}l@{}}0.623\\ $\pm$0.298\end{tabular} & \begin{tabular}[c]{@{}l@{}}0.731\\ $\pm$0.211\end{tabular} & {\begin{tabular}[c]{@{}l@{}}0.493\\ $\pm$0.268\end{tabular}} & \textbf{\begin{tabular}[c]{@{}l@{}}0.902\\ $\pm$0.123\end{tabular}} \\
    $\boldsymbol{25^{th} \%}$ & 0.242 & {\textbf{0.539}} & {0.283} & \multicolumn{1}{l|}{0.471} & 0.324 & 0.311 & 0.751 & \multicolumn{1}{l|}{0.593} & 0.329 & 0.632 & {0.262} & \textbf{0.878} \\
    \textbf{median} & 0.364 & {0.581} & {0.506} & \multicolumn{1}{l|}{\textbf{0.719}} & 0.527 & 0.582 & 0.821 & \multicolumn{1}{l|}{0.761} & 0.708 & 0.806 & {0.462} & \textbf{0.935} \\
    $\boldsymbol{75^{th} \%}$ & 0.659 & {0.675} & {0.768} & \multicolumn{1}{l|}{\textbf{0.888}} & 0.740 & 0.762 & 0.911 & \multicolumn{1}{l|}{0.898} & 0.898 & 0.865 & {0.714} & \textbf{0.970} \\ \cline{2-7} \cline{9-13} 
    \multicolumn{1}{c}{\textbf{}} & \multicolumn{6}{c}{\textbf{Cloth Fold Image}} & \textbf{} & \multicolumn{5}{c}{\textbf{Cloth Fold State}} \\ \cline{2-7} \cline{9-13} 
    $\boldsymbol{\mu \pm \sigma}$ & \begin{tabular}[c]{@{}l@{}}0.137\\ $\pm$0.096\end{tabular} & {\begin{tabular}[c]{@{}l@{}}-0.632\\ $\pm$1.264\end{tabular}} & {\begin{tabular}[c]{@{}l@{}}0.000\\ $\pm$0.000\end{tabular}} & \multicolumn{1}{l|}{\textbf{\begin{tabular}[c]{@{}l@{}}0.395\\ $\pm$0.318\end{tabular}}} & \begin{tabular}[c]{@{}l@{}}-0.530\\ $\pm$0.605\end{tabular} & \begin{tabular}[c]{@{}l@{}}-0.021\\ $\pm$0.237\end{tabular} & \begin{tabular}[c]{@{}l@{}}0.706\\ $\pm$0.159\end{tabular} & \multicolumn{1}{l|}{\begin{tabular}[c]{@{}l@{}}-0.277\\ $\pm$0.719\end{tabular}} & \begin{tabular}[c]{@{}l@{}}0.599\\ $\pm$0.246\end{tabular} & \begin{tabular}[c]{@{}l@{}}0.212\\ $\pm$0.431\end{tabular} & {\begin{tabular}[c]{@{}l@{}}-0.154\\ $\pm$0.491\end{tabular}} & \textbf{\begin{tabular}[c]{@{}l@{}}0.771\\ $\pm$0.117\end{tabular}} \\
    $\boldsymbol{25^{th} \%}$ & \textbf{0.090} & {0.000} & {0.000} & \multicolumn{1}{l|}{0.000} & -0.789 & -0.001 & 0.637 & \multicolumn{1}{l|}{-0.538} & 0.506 & -0.083 & {-0.255} & \textbf{0.720} \\
    \textbf{median} & 0.159 & {0.000} & {0.000} & \multicolumn{1}{l|}{\textbf{0.493}} & -0.350 & 0.000 & 0.726 & \multicolumn{1}{l|}{-0.145} & 0.669 & 0.002 & {-0.025} & \textbf{0.776} \\
    $\boldsymbol{75^{th} \%}$ & 0.210 & {0.000} & {0.000} & \multicolumn{1}{l|}{\textbf{0.668}} & -0.039 & 0.000 & 0.815 & \multicolumn{1}{l|}{0.038} & 0.774 & 0.661 & {0.063} & \textbf{0.846} \\ \cline{2-7}
    \multicolumn{1}{c}{\textbf{}} & \multicolumn{6}{c}{\textbf{Cloth Fold Diagonal Pinned Image}} & \multicolumn{1}{c}{\cellcolor[HTML]{EFEFEF}\textbf{}} & \multicolumn{5}{c}{} \\ \cline{2-7}
    $\boldsymbol{\mu \pm \sigma}$ & \begin{tabular}[c]{@{}l@{}}0.570\\ $\pm$0.349\end{tabular} & {\begin{tabular}[c]{@{}l@{}}0.379\\ $\pm$0.249\end{tabular}} & {\begin{tabular}[c]{@{}l@{}}0.521\\ $\pm$0.080\end{tabular}} & \multicolumn{1}{l|}{\textbf{\begin{tabular}[c]{@{}l@{}}0.895\\ $\pm$0.010\end{tabular}}} & \begin{tabular}[c]{@{}l@{}}0.775\\ $\pm$0.035\end{tabular} & \begin{tabular}[c]{@{}l@{}}0.679\\ $\pm$0.044\end{tabular} & \begin{tabular}[c]{@{}l@{}}0.906\\ $\pm$0.009\end{tabular} & \multicolumn{5}{c}{} \\
    $\boldsymbol{25^{th} \%}$ & 0.276 & {0.448} & {0.454} & \multicolumn{1}{l|}{\textbf{0.892}} & 0.763 & 0.657 & 0.898 & \multicolumn{5}{c}{} \\
    \textbf{median} & 0.743 & {0.454} & {0.460} & \multicolumn{1}{l|}{\textbf{0.896}} & 0.775 & 0.678 & 0.905 & \multicolumn{5}{c}{} \\
    $\boldsymbol{75^{th} \%}$ & 0.896 & {0.461} & {0.614} & \multicolumn{1}{l|}{\textbf{0.899}} & 0.785 & 0.695 & 0.914 & \multicolumn{5}{c}{} \\ \cline{2-7}
    \multicolumn{1}{c}{\textbf{}} & \multicolumn{6}{c}{\textbf{Cloth Fold Diagonal Unpinned Image}} & \multicolumn{1}{c}{\cellcolor[HTML]{EFEFEF}\textbf{}} & \multicolumn{5}{c}{} \\ \cline{2-7}
    $\boldsymbol{\mu \pm \sigma}$ & \begin{tabular}[c]{@{}l@{}}0.905\\ $\pm$0.009\end{tabular} & {\begin{tabular}[c]{@{}l@{}}0.309\\ $\pm$0.255\end{tabular}} & {\begin{tabular}[c]{@{}l@{}}0.546\\ $\pm$0.065\end{tabular}} & \multicolumn{1}{l|}{\textbf{\begin{tabular}[c]{@{}l@{}}0.940\\ $\pm$0.035\end{tabular}}} & \begin{tabular}[c]{@{}l@{}}0.835\\ $\pm$0.047\end{tabular} & \begin{tabular}[c]{@{}l@{}}0.789\\ $\pm$0.036\end{tabular} & \begin{tabular}[c]{@{}l@{}}0.927\\ $\pm$0.011\end{tabular} & \multicolumn{5}{c}{} \\
    $\boldsymbol{25^{th} \%}$ & 0.903 & {0.000} & {0.499} & \multicolumn{1}{l|}{\textbf{0.912}} & 0.811 & 0.771 & 0.918 & \multicolumn{5}{c}{} \\
    \textbf{median} & 0.907 & {0.500} & {0.505} & \multicolumn{1}{l|}{\textbf{0.951}} & 0.846 & 0.784 & 0.930 & \multicolumn{5}{c}{} \\
    $\boldsymbol{75^{th} \%}$ & 0.911 & {0.520} & {0.582} & \multicolumn{1}{l|}{\textbf{0.975}} & 0.870 & 0.811 & 0.937 & \multicolumn{5}{c}{
    \multirow{-10}{*}[2em]{
    \fbox{
        \begin{minipage}{0.33\textwidth}
                \centering
                \includegraphics[width=\linewidth]{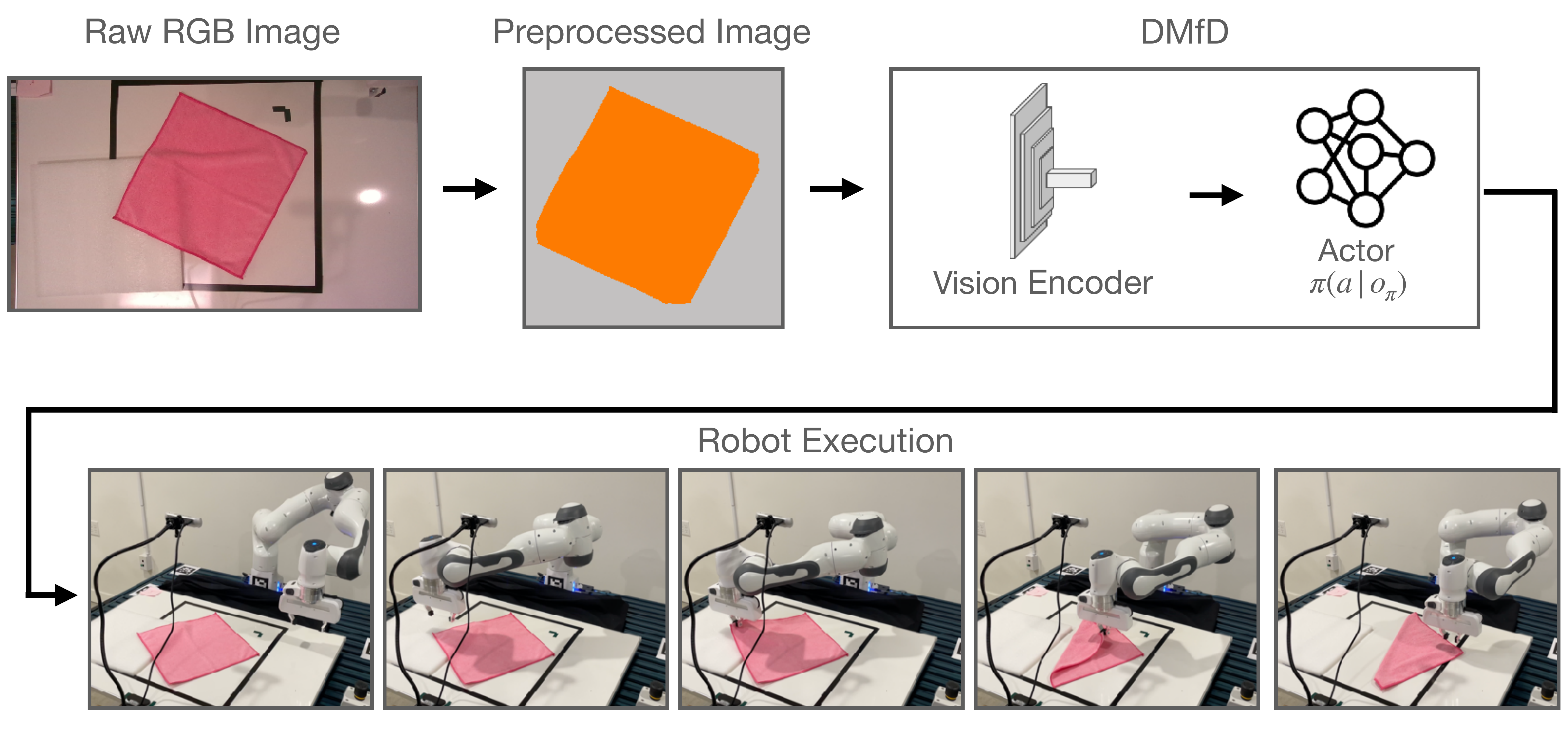}
                \caption*{\resubmit{Pipeline for real robot experiments.}}
                \label{fig:real-robot-pipeline}
        \end{minipage}
        }
    }
    }
    \end{tabular}
    \caption{\resubmit{\textbf{End of training performance comparison.} Performance metric with normalized performance $\hat{p}(H)$, using the models at the end of training. Models are obtained at the end of training for each method (total of 5 seeds). We show the mean, variance, median, $25^{th}$ and $75^{th}$ percentiles of performance over 100 evaluations of each method. A vertical line is drawn to differentiate between methods that use expert demonstrations, and methods that do not. Experts are oracles with state information, shown in grey. The inset figure shows the pipeline for real robot experiments.}}
    \label{tab:sota-comparisons-all}
    \end{table*}
    
\begin{figure*}%
\begin{subfigure}{0.7\columnwidth}
  \centering
  \includegraphics[width=\linewidth]{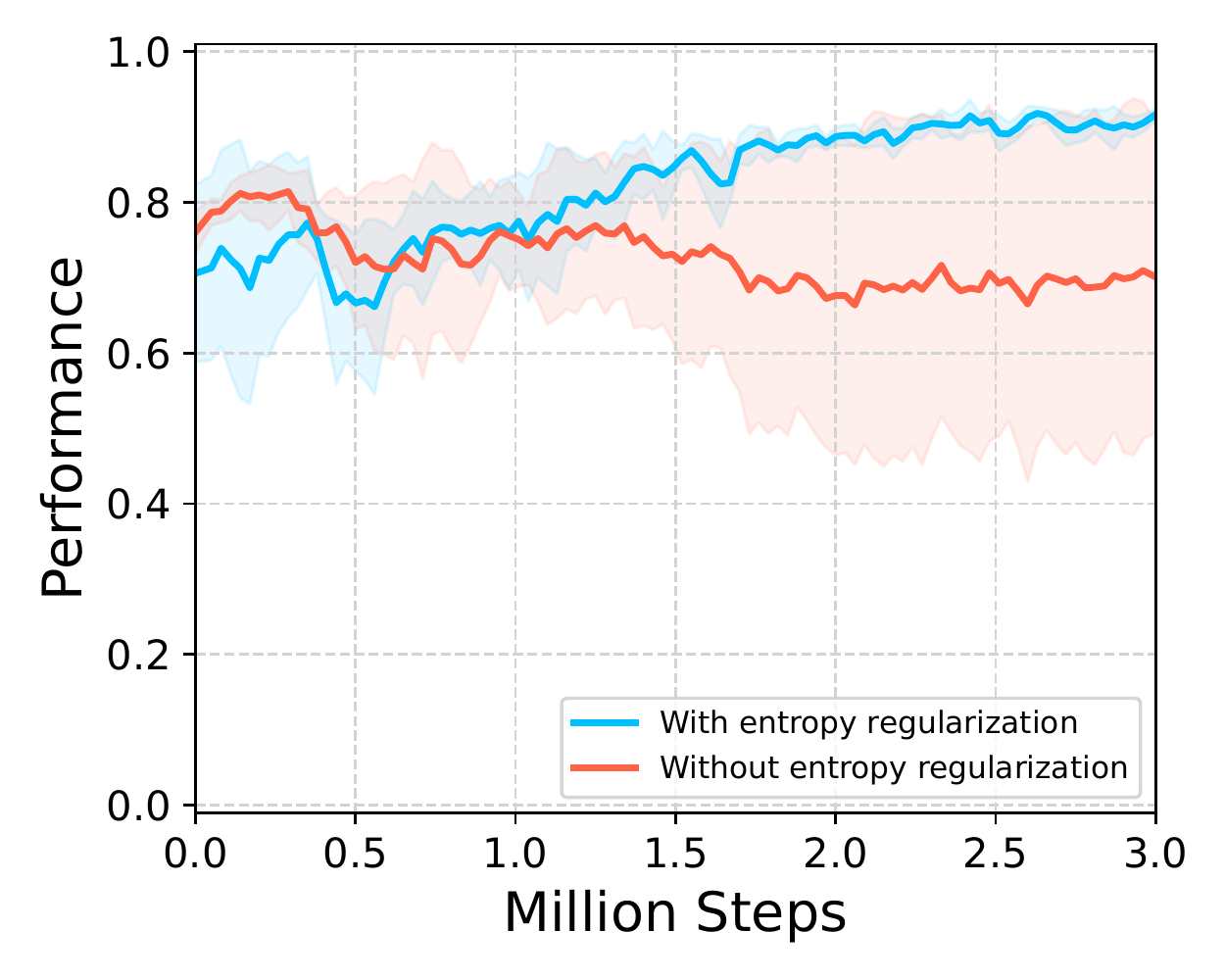}
  \caption{Entropy regularization}
  \label{fig:abl-sac-loss}
\end{subfigure}
\begin{subfigure}{0.7\columnwidth}
  \centering
  \includegraphics[width=\linewidth]{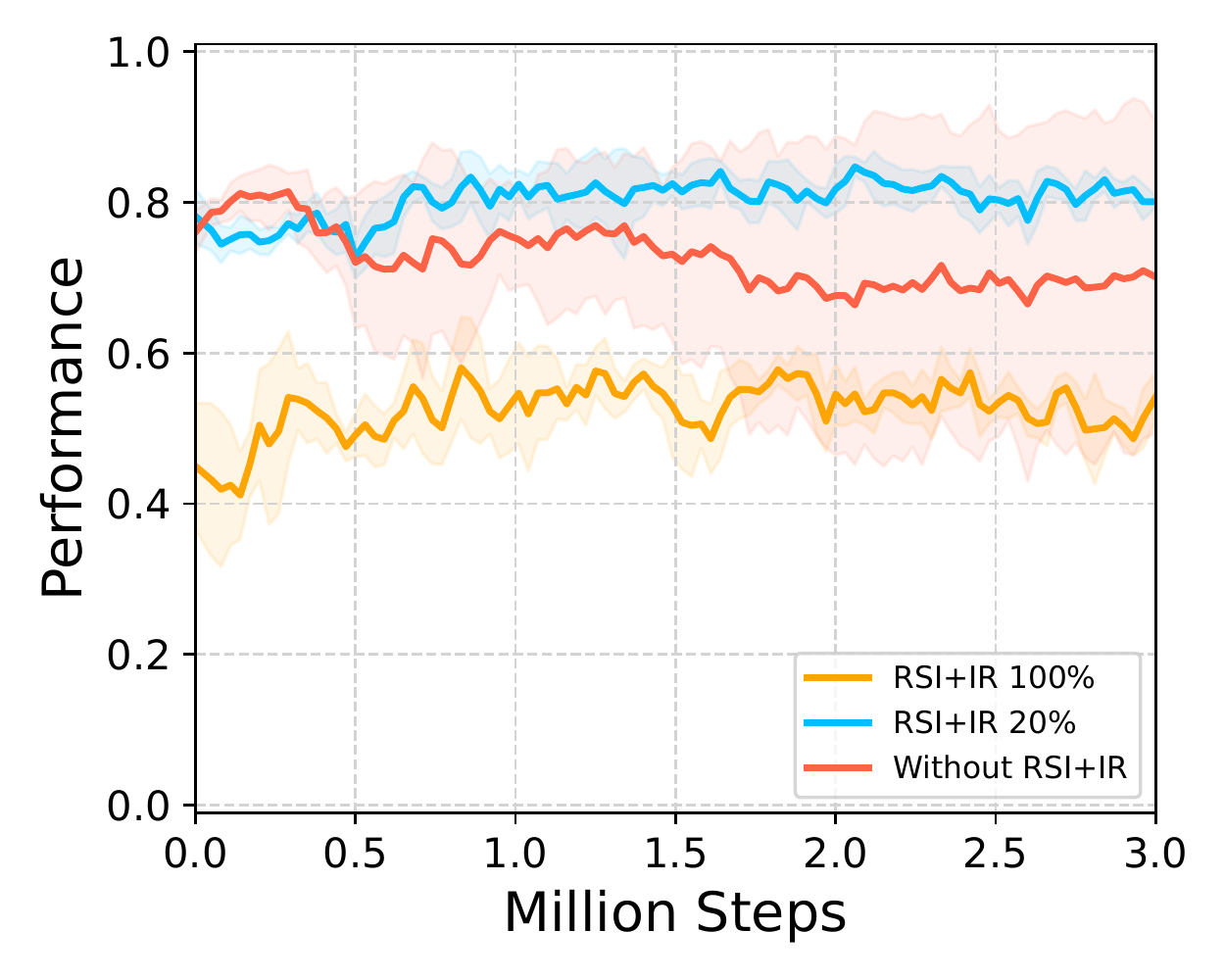}
  \caption{\resubmit{RSI, State-based observations}}
  \label{fig:abl-rsi-state}
\end{subfigure}
\begin{subfigure}{0.7\columnwidth}
  \centering
  \includegraphics[width=\linewidth]{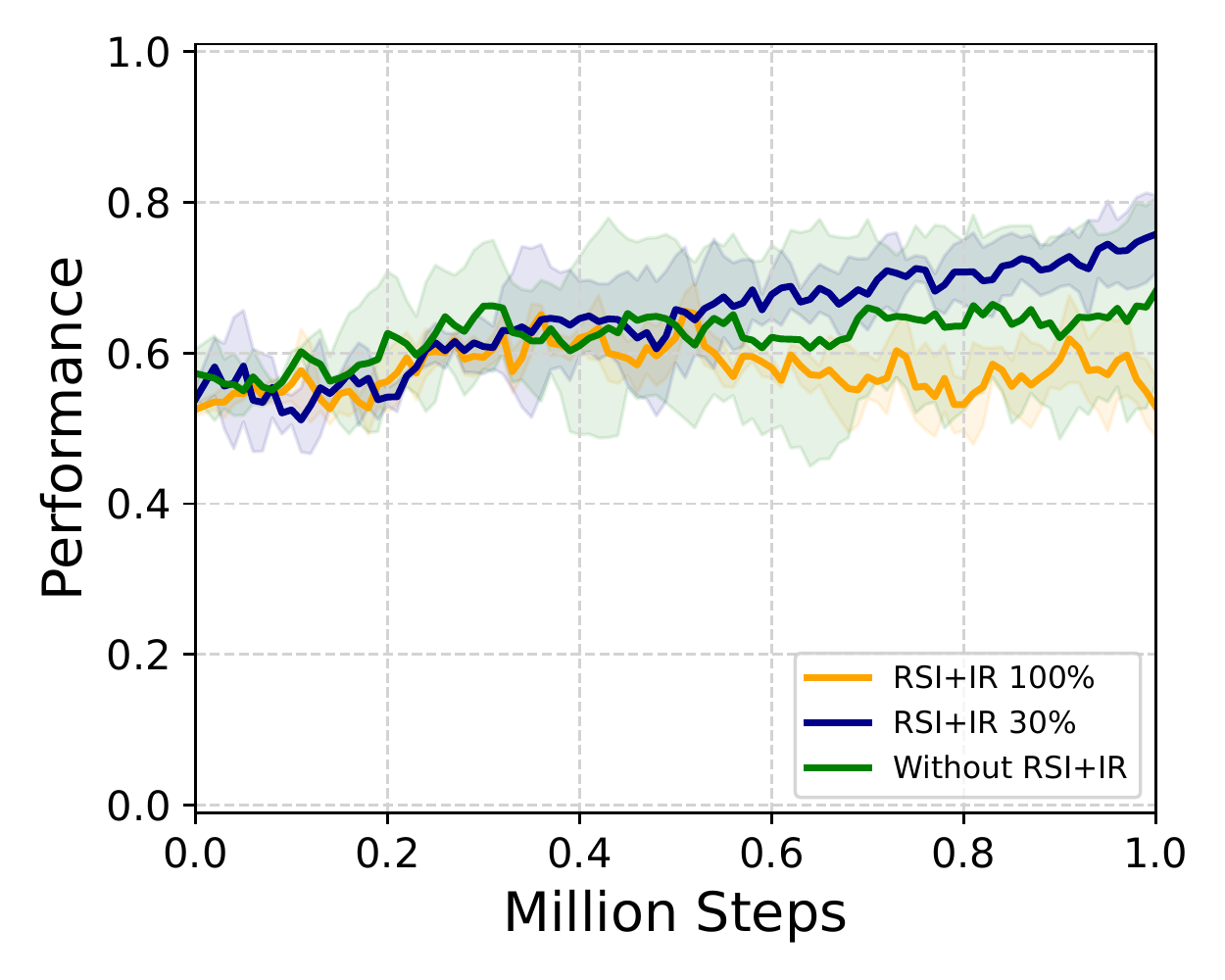}
  \caption{\resubmit{RSI, Image-based observations}}
  \label{fig:abl-rsi-image}
\end{subfigure}
\begin{subfigure}{0.7\columnwidth}
  \centering
  \includegraphics[width=\linewidth]{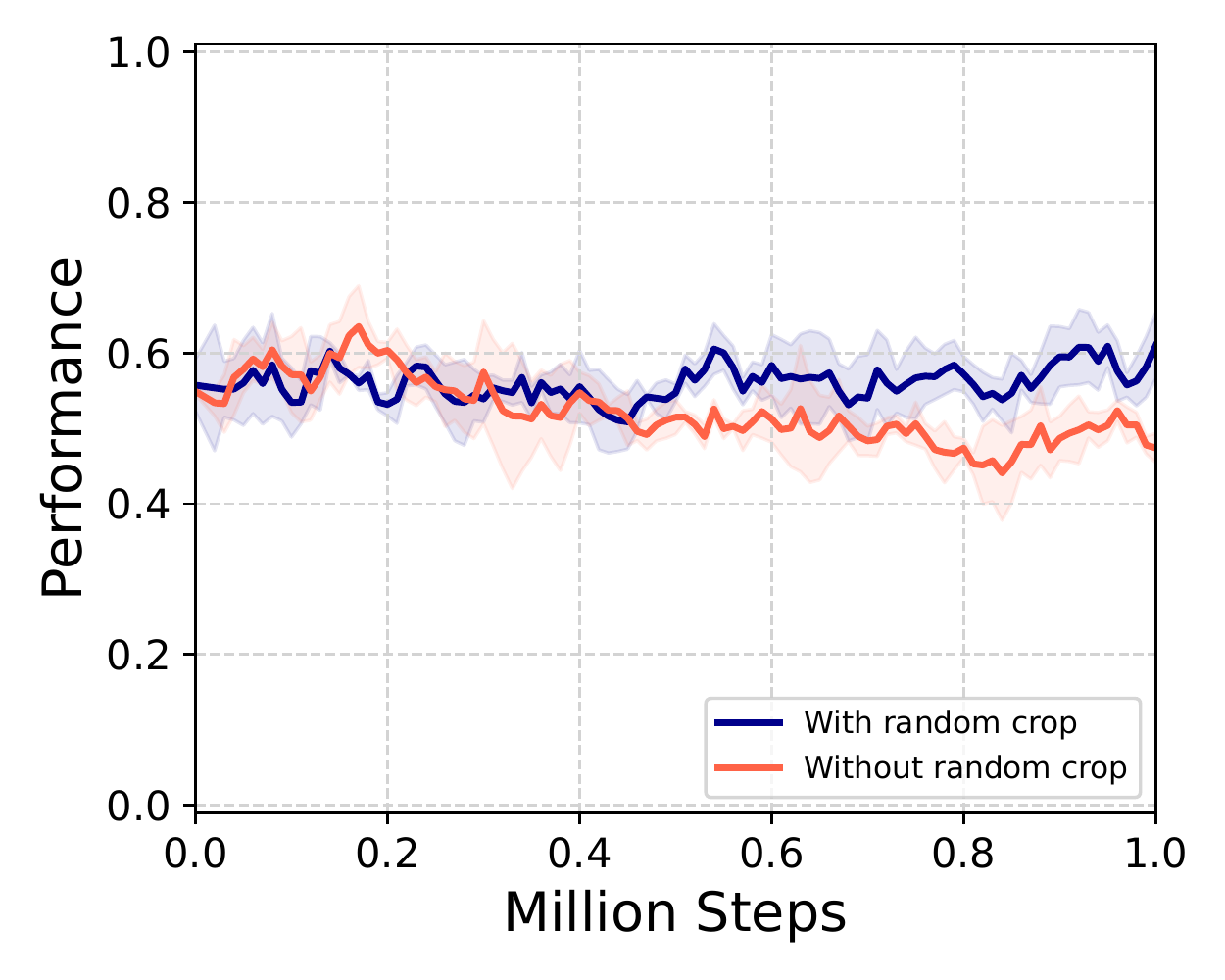}
  \caption{Random Crop}
  \label{fig:abl-random-crop}
\end{subfigure}
\begin{subfigure}{0.7\columnwidth}
  \centering
  \includegraphics[width=\linewidth]{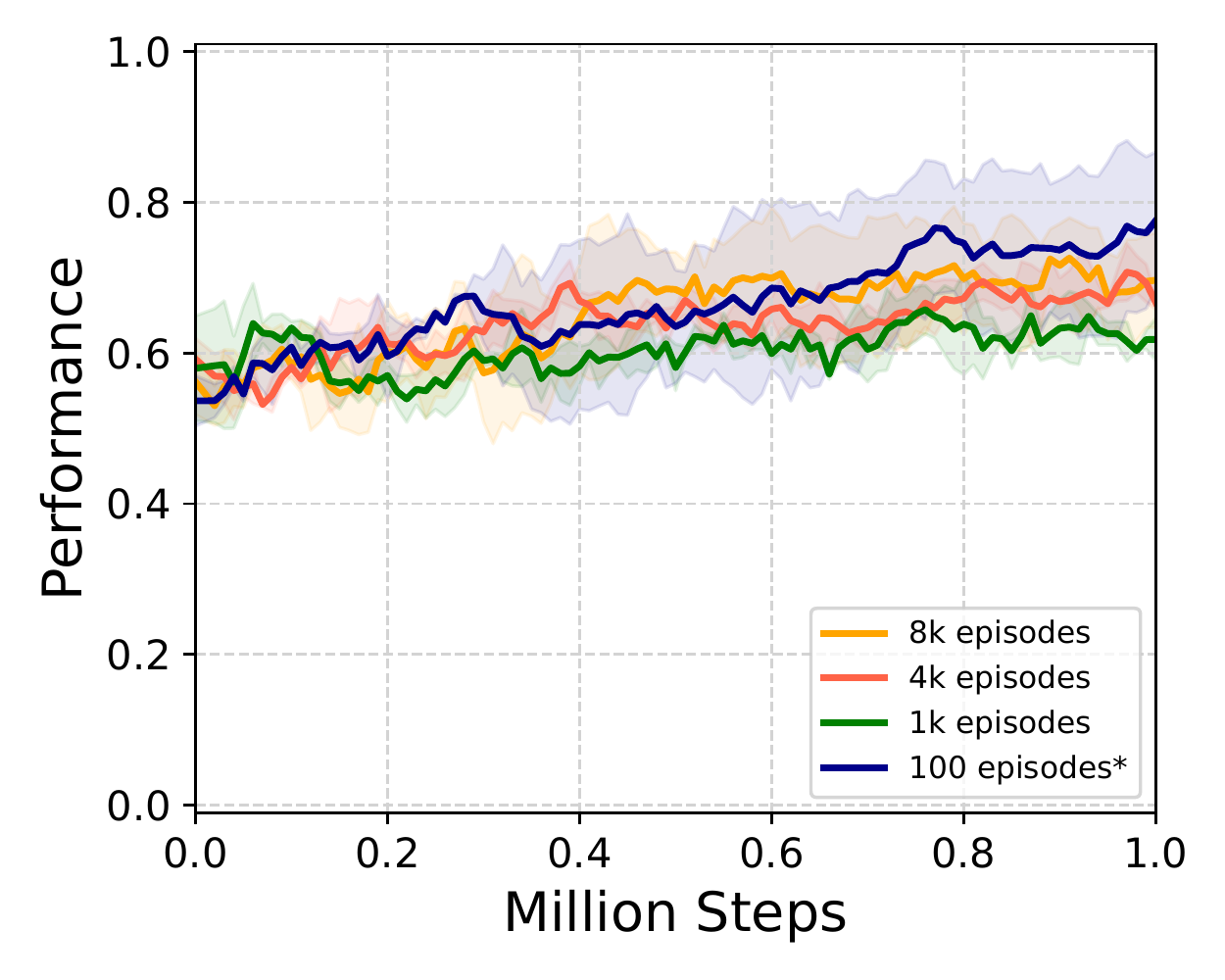}
  \caption{\resubmit{\cameraready{Expert Dataset Size}}}
  \label{fig:abl-dataset-size}
\end{subfigure}
\begin{subfigure}{0.7\columnwidth}
  \centering
  \includegraphics[width=\linewidth]{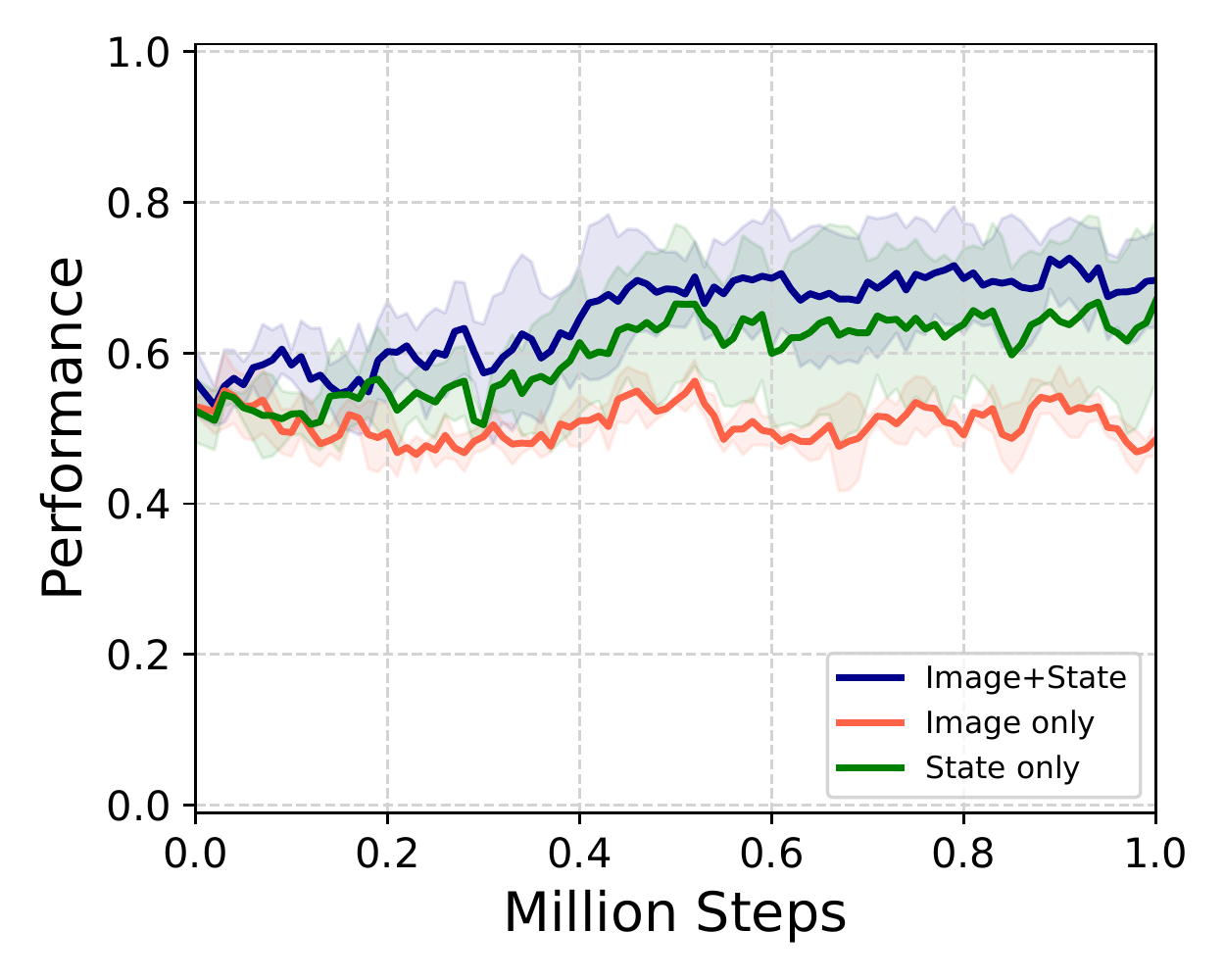}
  \caption{\resubmit{Critics inputs}}
  \label{fig:abl-critic}
\end{subfigure}

\caption{\resubmit{\textbf{Ablation studies.} Ablations were performed on the Straighten Rope environment, to verify the necessity for each feature used. 
State-based DMfD is shown in light blue, and image-based DMfD is in dark blue. 
Entropy regularization~\autoref{fig:abl-sac-loss} and one ablation for reference state-initialization~\autoref{fig:abl-rsi-state} were run on the state-based environment.
The other ablations (\autoref{fig:abl-rsi-image}, \autoref{fig:abl-random-crop}, \autoref{fig:abl-dataset-size}, and \autoref{fig:abl-critic}) require an image-based environment. 
We plot the mean $\mu$ of the curves as a solid line, and shade one standard deviation ($\mu \pm \sigma$). 
Detailed discussion of these features is in \autoref{sec:discussion}.
In \autoref{fig:abl-dataset-size}, `100 episodes$^*$' refers to 100 episodes of data copied 80 times to mimic the buffer of 8000 episodes without actually creating as many expert demonstrations.}}
\label{fig:ablations-all}
\end{figure*}

\resubmit{We compare our method with these Non-LfD baselines:}
\begin{itemize}
    \item \textbf{SAC}: A SOTA off-policy actor-critic RL algorithm.
    \item \textbf{SAC-CURL}: A SOTA off-policy image-based RL algorithm using contrastive learning~\cite{Laskin2020CuRL}.
    \item \textbf{SAC-DrQ}: A SOTA off-policy image-based RL algorithm using data augmentations and regularized Q-function~\cite{Kostrikov2020DrQ}.
\end{itemize}

We also compare with these LfD baselines:
\begin{itemize}
     \item \textbf{AWAC}: A SOTA off-policy RL algorithm that learns from offline data followed by online fine-tuning~\cite{nair2020AWAC}.
    \item \textbf{BC-State}: A behavior cloning policy trained on state-action pairs~\cite{Torabi2018BehavorialCloning}.
    \item \textbf{SAC-LfD}: SAC with pre-populated expert data in the replay buffer.
    \item \textbf{BC-Image}: A behavior cloning policy trained on image-action pairs~\cite{Torabi2018BehavorialCloning}.
    \item \textbf{SAC-BC}: SAC with initialized actor networks from pre-trained BC-Image on expert demonstrations.

\end{itemize}
\color{black} %

\resubmit{We use Softgym's implementations and hyperparameters for baselines where applicable, taken from the official implementations of the algorithms cited.
We did not include the PlaNet~\cite{Hafner2019PlaNet} baseline as it did not beat the other image-based baselines. \autoref{fig:sota-comparisons-all} shows training curves and \autoref{tab:sota-comparisons-all} shows the comparison at the end of training. DMfD outperforms all baselines. For both state- and image-based environments as the tasks get more difficult, DMfD outperforms baselines by higher margins. A detailed discussion is in~\autoref{sec:discussion}.} 

\subsection{\resubmit{Real Robot Experiments}}

\noindent \resubmit{\textbf{Setup:} We use the DMfD model trained in simulation to perform the Cloth Fold Diagonal Unpinned task on a Franka Emika Panda robot arm and the default gripper.
An Intel RealSense camera is used to capture RGB images of a top-down view of the cloth.
To obtain real-world images that resemble simulated images, we center crop the original RGB images from the camera, segment the cloth from the background, and fill the cloth and the background with colors from the simulated cloth and table, ensuring robustness to different colors of the cloth and background in the real-world setup. Our method does not require any training or fine-tuning in the physical setting.}

\noindent \resubmit{\textbf{Results:} We evaluate our method on ten rollouts. Each rollout has a different cloth orientation (ranging from -19\textdegree ~to +25\textdegree). 
We use a checkpoint from 60,000 environment steps in simulation to initialize the actor of our agent. In simulation, our policy has a mean accuracy of 91.28\% over ten rollouts. On the real robot, we obtain a mean accuracy of 85.58\%.}

\subsection{Ablations}
\label{sec:ablations}

We test our ablations over 3 seeds each, and plot the mean and variance of performance during training. 
We run these ablations in the Straighten Rope environment, with state- and image-based observations as applicable.

\noindent \textbf{Entropy Regularization:} \autoref{fig:abl-sac-loss} \resubmit{shows that using entropy regularization enables the agent to explore the environment further, surpassing its initial performances of learning from the expert data in the replay buffer.}
We see high variance in the baseline, indicating less robustness to randomness (e.g., seed, task variants, etc.) and unstable training performance. 

\noindent \textbf{\resubmit{Probabilistic Reference State Initialization~(RSI):}} \autoref{fig:abl-rsi-state} and \autoref{fig:abl-rsi-image} show ablations for using RSI. \resubmit{With the default configuration of RSI (RSI+IR 100\%), the agent shows worse performance than not using RSI. 
In other words, simply applying RSI in deformable object manipulation may lead to poor results due to constantly resetting the agent to the predefined states, which prevents the agent from freely exploring the environment. 
However, the agent can benefit from expert demonstrations without limiting exploration by invoking RSI probabilistically.}
\label{sec:abl-rsi}

\noindent \textbf{Random Crops of Image Observations:}  \resubmit{\autoref{fig:abl-random-crop} shows that using random crops as an augmentation technique improves performance. This confirms that employing random crops stabilizes visual RL training which would otherwise overfit.}

\noindent \textbf{Number of Expert Trajectories:} \label{sec:abl-prepopulatebuffer} We estimate how much expert data is optimal for our agent (\autoref{fig:abl-dataset-size}). 
Using 1000 expert episodes is noticeably worse than 4000 episodes.
However, the difference between 4000 and 8000 episodes is small, indicating that the marginal utility of adding additional expert trajectories reduces with the number of trajectories.
\resubmit{To test the sample efficiency of DMfD, we propose to use only 100 episodes of data but duplicate them to fill the replay buffer. 
As shown, it is possible to achieve similar performance as by using larger amounts of data, but we have found it to be less robust to environment variations and training seeds. 
We conclude it is best to use as many expert demonstrations as possible when expert demonstrations are easily obtainable. 
However, when they are not readily available, duplicating expert data to fill up the replay buffer is a viable way to learn from expert demonstrations.}

\resubmit{
\noindent \textbf{Critic Inputs:} \label{sec:abl-critic} In \autoref{fig:abl-critic}, we examine the effect of different types of inputs to the critics. 
Having states information as input to the critics is essential in obtaining higher performance. 
This is because states contain valuable information vital to the tasks but not readily interpretable in images (e.g., the true coordinates of cloth corners).
However, the addition of images to states has the best performance (~\autoref{fig:abl-critic}), likely because the critics are able to see what the actor sees, and can provide a better guiding value estimate.
}

\subsection{Discussion}
\label{sec:discussion}

Compared to experts, \autoref{tab:sota-comparisons-all} shows that our state-based agent beats the expert in both state environments. 
The image-based solutions are comparable to the expert at best, as they do not have privileged state information. When comparing with baselines, we see that the performance gap between our method and baselines increases with respect to the difficulty of the task. In easier tasks, our method's capabilities are not fully utilized. We observe this for both state- and image-based environments. For example, in a harder task like Cloth Fold Image~\autoref{fig:sota-cloth-fold-image}, the baseline methods are at or below 0 performance at the end of training. 

Because we use expert data in multiple ways, our state-based method outperforms SAC, a baseline that does not use expert data. 
The lack of expert data \resubmit{severely} affects the performance of SAC on the difficult state-based task, Cloth Fold. 
The benefits of using expert data, in all environments, are shown in \autoref{fig:abl-rsi-state}, \autoref{fig:abl-rsi-image} and \autoref{fig:abl-dataset-size}. Moreover, given a pre-populated replay buffer, we can think of RSI giving DMfD an extra boost essentially for \resubmit{`free'} (since we reuse the same expert data).
Conversely, AWAC achieves better performance on difficult tasks with the help of expert data. 
However, a lack of entropy regularization means that it is more prone to reaching a local optimum during training. 
This can be seen in the higher variance than DMfD during training, indicating lower robustness to randomness. %
In fact, in the Straighten Rope Image experiment, this high variance after 1M steps eventually leads to a deterioration in performance.

Image-based environments are harder and this is where DMfD outperforms the baselines even further. \resubmit{In image-based environments, LfD baselines outperform non-LfD baselines in the Straighten Rope, Cloth Fold, and Cloth Fold Diagonal Unpinned environments. However, non-LfD methods have more consistent performance than LfD methods. This implies that LfD baselines are not as robust as the non-LfD methods, and they may require more sophisticated solutions for consistently better performance. In other words, designing a robust LfD method in these environments is nontrivial. As shown in \autoref{fig:sota-comparisons-all} and \autoref{tab:sota-comparisons-all}, DMfD consistently outperforms all baselines. It is adept at learning these challenging tasks while leveraging expert demonstrations.} The experiments provide strong evidence that DMfD is consistently equal or better performant than the baselines across all environments, while being robust to noise. 

\section{Conclusion}
\label{sec:conclusion}
We describe a new reinforcement learning-based method - Deformable Manipulation from Demonstrations (DMfD) - that leverages expert demonstrations and \resubmit{outperforms state-of-the-art Learning from Demonstration (LfD) methods} for representative manipulation tasks on 1D (rope) and 2D (cloths) deformable objects. 
For both state-based and image-based inputs, DMfD effectively leverages expert demonstrations as follows: \resubmit{1. we pre-populate the replay buffer with expert trajectories before training, 2. during training, we improve on the standard advantage-weighted loss by adding an exploration term (and extending it to image-based inputs), and 3. during experience collection we improve on reference state initialization by using it probabilistically.} 
For image-based inputs, we use an asymmetric actor-critic architecture, where the actor acts based solely on environment images while the critics learn from both image and state information. 
To make our policy more robust to different variations of the environments, we applied random cropping to sampled images during the actor-critic updates. We demonstrate the effectiveness of DMfD on two challenging deformable object manipulation tasks from the SoftGym suite. We also create two new challenging environments for folding a 2D cloth using image-based observations, and set a performance benchmark for them. 
\resubmit{We show a consistent and noticeable performance improvement over baselines in state-based environments (up to 12.9\% on median) and an even higher improvement on tougher image-based environments (up to 33.44\% on median).} 
We also observe comparable or lower variance than the baselines, indicating higher robustness to noise. \resubmit{To validate the feasibility of DMfD in real-world settings, we conducted real robot experiments and achieved a minimal sim2real gap ($\sim$6\%)} \cameraready{in normalized performance.}

\bibliographystyle{IEEEtran}
\bibliography{references}

\end{document}